\documentclass[preprint,12pt]{elsarticle}

\usepackage{amsmath,amsfonts}
\usepackage{algorithmic}
\usepackage{algorithm}
\usepackage{array}
\usepackage{textcomp}
\usepackage{stfloats}
\usepackage{url}
\usepackage{verbatim}
\usepackage{graphicx}
\usepackage[mathscr]{eucal}
\usepackage{tikz}

\usepackage{shadowtext}
\usepackage{filecontents}
\shadowoffset{2pt}

\usepackage{subcaption}
\usepackage{cleveref}
\usepackage{multirow}

\newcommand{\ie}{\textit{i}.\textit{e}.}

\usepackage{pifont}
\usepackage[perpage,symbol*]{footmisc}
\DefineFNsymbols{circled}{{\ding{192}}{\ding{193}}{\ding{194}}
{\ding{195}}{\ding{196}}{\ding{197}}{\ding{198}}{\ding{199}}{\ding{200}}{\ding{201}}}
\setfnsymbol{circled}

\newcommand\myfootnotestyle[1]{\ifcase#1 \or \ding{182}\or \ding{183}\or
\ding{184}\or \ding{185}\or \ding{186}\or \ding{187}%
\or \ding{188}\or \ding{189}\or \ding{190}\or \ding{191}\else *\fi\relax}

\journal{Neural Networks}

\begin{document}

\begin{frontmatter}



\title{Attacking Cooperative Multi-Agent Reinforcement Learning by Adversarial Minority Influence} 


\author{Simin Li\textsuperscript{a, e}, Jun Guo\textsuperscript{a}, Jingqiao Xiu\textsuperscript{a}, Yuwei Zheng\textsuperscript{a}, Pu Feng\textsuperscript{a}, Xin Yu\textsuperscript{a}, Jiakai Wang\textsuperscript{b}, Aishan Liu\textsuperscript{a}, Yaodong Yang\textsuperscript{d}, Bo An\textsuperscript{e}, Wenjun Wu\textsuperscript{a}, Xianglong Liu\textsuperscript{a, b, c} \footnote{Corresponding author: Xianglong Liu. Email address: xlliu@buaa.edu.cn.}}
\affiliation{organization={State Key Lab of Software Development Environment, Beihang University},
            city={Beijing},
            country={China}}
\affiliation{organization={Zhongguancun Laboratory},
            city={Beijing},
            country={China}}
\affiliation{organization={Institute of Data Space, Hefei Comprehensive National Science Center},
            city={Hefei},
            country={China}}
\affiliation{organization={Institute of Artificial Intelligence, Peking University},
            city={Beijing},
            country={China}}
\affiliation{organization={Nanyang Technological University},
            country={Singapore}}

\begin{abstract}
This study probes the vulnerabilities of cooperative multi-agent reinforcement learning (c-MARL) under adversarial attacks, a critical determinant of c-MARL's worst-case performance prior to real-world implementation. Current observation-based attacks, constrained by white-box assumptions, overlook c-MARL's complex \emph{multi-agent} interactions and \emph{cooperative} objectives, resulting in impractical and limited attack capabilities. To address these shortcomes, we propose \emph{Adversarial Minority Influence} (AMI), a practical and strong for c-MARL. AMI is a practical black-box attack and can be launched without knowing victim parameters. AMI is also strong by considering the complex \emph{multi-agent} interaction and the \emph{cooperative} goal of agents, enabling a single adversarial agent to \emph{unilaterally} misleads majority victims to form \emph{targeted} worst-case cooperation. This mirrors minority influence phenomena in social psychology. To achieve maximum deviation in victim policies under complex agent-wise interactions, our \emph{unilateral} attack aims to characterize and maximize the impact of the adversary on the victims. This is achieved by adapting a unilateral agent-wise relation metric derived from mutual information, thereby mitigating the adverse effects of victim influence on the adversary. To lead the victims into a jointly detrimental scenario, our \emph{targeted} attack deceives victims into a long-term, cooperatively harmful situation by guiding each victim towards a specific target, determined through a trial-and-error process executed by a reinforcement learning agent. Through AMI, we achieve the first successful attack against real-world robot swarms and effectively fool agents in simulated environments into collectively worst-case scenarios, including Starcraft II and Multi-agent Mujoco. The source code and demonstrations can be found at: \url{https://github.com/DIG-Beihang/AMI}.
\end{abstract}

\begin{graphicalabstract}
\begin{figure}[t]
\centering
\includegraphics[scale=0.45]{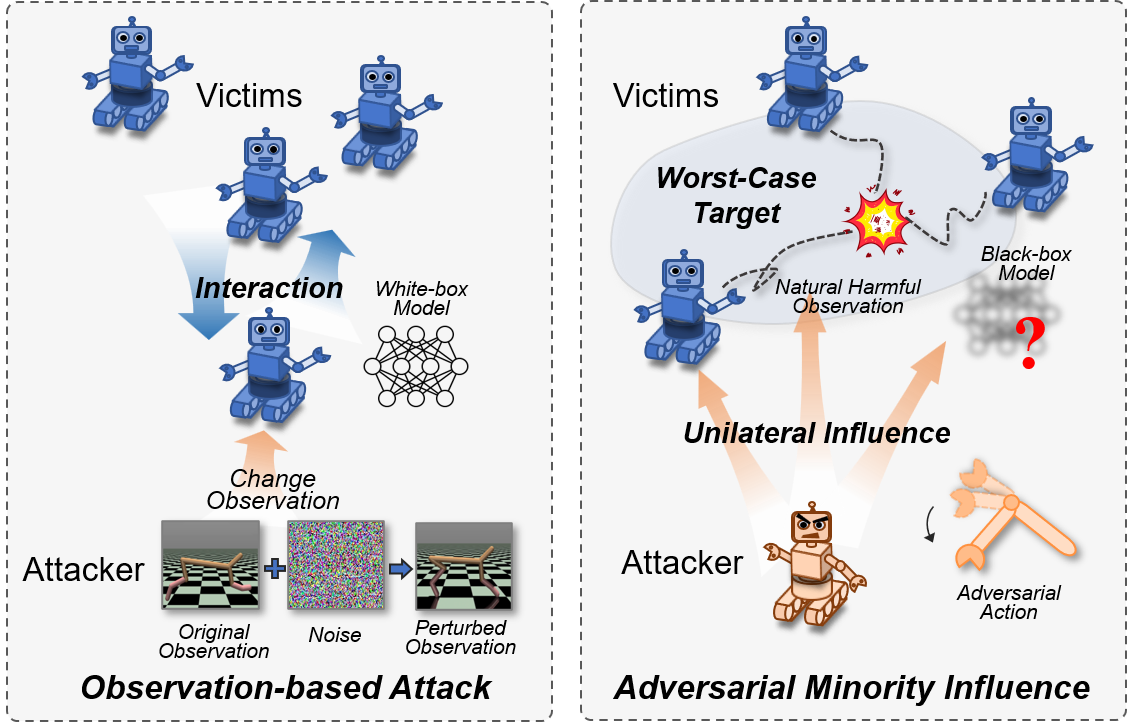}
\end{figure}
\end{graphicalabstract}

\begin{highlights}
\item We develop AMI, a strong and practical attack towards c-MARL, which leverages the intricate influence among agents and the cooperative nature of victims in c-MARL.
\item We introduce the \emph{unilateral influence filter} and \emph{targeted adversarial oracle} to optimize the deviation of victim policies and deceive victims into suboptimal cooperation, thereby ensuring a potent attack capability.
\item AMI achieves the first successful attack against real world robot swarms and effectively fool agents in simulation environments into collectively worst-case scenarios, including StarCraft II and Multi-agent Mujoco.
\end{highlights}


\end{frontmatter}



\section{Introduction}

Cooperative multi-agent reinforcement learning (c-MARL) involves the coordination of multiple agents to maximize their shared objective in an environment \cite{rashid2018qmix, lowe2017maddpg, yu2021mappo, ji2023ai, ji2024aligner, ji2024beavertails, ji2024language}. Applications of c-MARL includes cooperative gaming \cite{samvelyan2019smac}, traffic signal management \cite{chu2019traffic}, distributed resource allocation \cite{zhang2009resourceallocation3}, and cooperative swarm control \cite{peng2021facmac, huttenrauch2019swarm, yun2022multiuav, yu2021mappo, yu2023esp, yu2021swarm}.

\begin{figure}[t]
\centering
\includegraphics[scale=0.45]{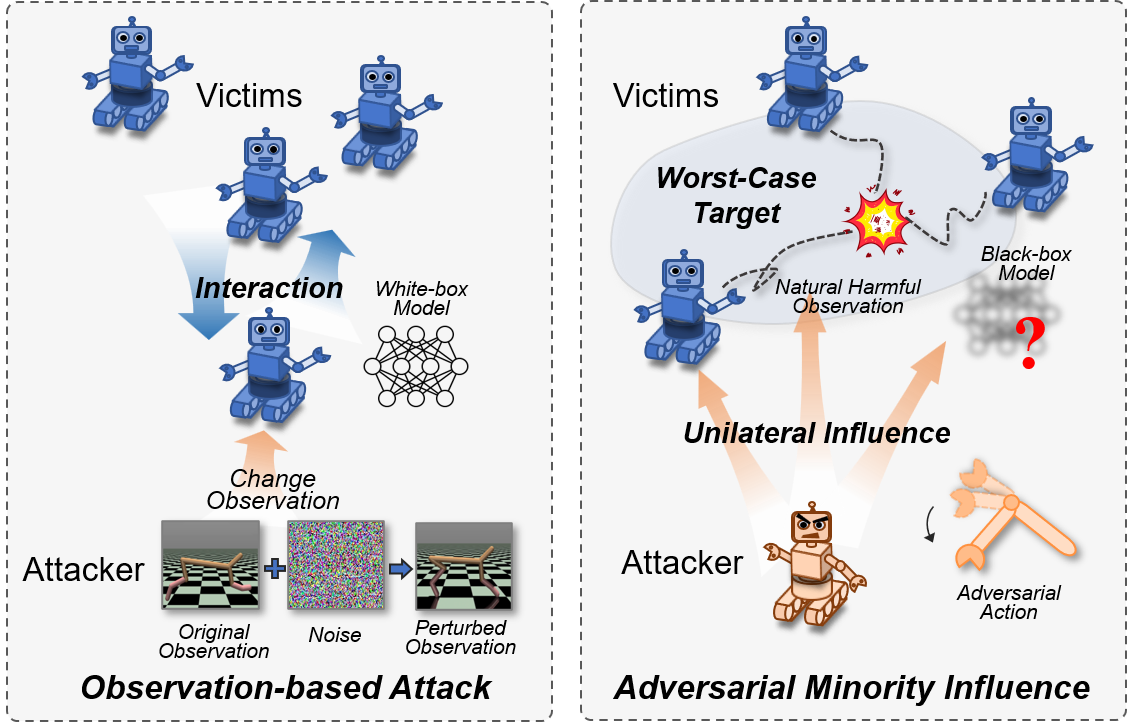}
\caption{While observation-based attack requires white-box assess to victim and manipulates agent observation directly, our adversarial minority influence attack is a black-box, policy-based attack that leverages one minority attacker to unilaterally influence majority victims towards a jointly worst target.}
\label{motivation}
\vspace{-0.2in}
\end{figure}

While c-MARL has achieved notable success, research has exposed the vulnerability of c-MARL agents to observation-based adversarial attacks \cite{lin2020jieyulins&p, zan2023obsbased}, wherein adversaries introduce perturbations to an agent's observation, causing it to execute suboptimal actions. Given the interrelated nature of victim actions for cooperation, other c-MARL agents may become disoriented and being non-robust (see Fig. \ref{motivation}). As c-MARL algorithms frequently feature in security-sensitive applications, assessing their worst-case performance against potential adversarial interference is crucial before real world deployment. However, observation-based attacks against c-MARL depend on white-box access to victim and complete control of agent observations, rendering them highly impractical. For instance, in autonomous driving scenarios, it can be prohibitively hard for attackers to access the architectures, weights, and gradients utilized by a vehicle or to introduce arbitrary pixel-wise manipulations to camera input at each timestep \cite{lin2020jieyulins&p, huang2017statebased1, zhang2020samdp}.

In this paper, we take a practical and black-box alternative by conducting a policy-based adversarial attack (\emph{i.e.}, adversarial policy) \cite{gleave2019iclr2020advpolicy, wu2021usenix, guo2021icml2021} to assess the robustness of c-MARL. Unlike direct observation manipulation, policy-based attacks perturb observations in a natural manner by incorporating an adversarial agent within the c-MARL environment—a feasible approach in numerous c-MARL applications. For instance, adversaries can legitimately participate in distributed resource allocation clusters as individual workers \cite{wu2018resourceallocation1, zhang2009resourceallocation3} or control their own vehicles while influencing other vehicles in autonomous driving \cite{shalev2016auto1, bhalla2020auto2}. By interacting with the environment, the adversary learns an adversarial policy that directs it to execute physically plausible actions, which in turn adversely influence the observations of victim agents.

Although policy-based attacks have been investigated in two-agent competitive games \cite{gleave2019iclr2020advpolicy, wu2021usenix, guo2021icml2021}, these studies have neglected two crucial challenges in c-MARL, resulting in diminished attack efficacy. Ideally, the adversary should \emph{influence} victims toward a \emph{cooperatively} inferior policy. This gives rise to two issues: (1) The \emph{influence} problem, wherein all agents mutually influence one another in c-MARL; consequently, attacking a single victim impacts the policies of other victims as well. In the face of such intricate agent-wise influence, it becomes difficult for the adversary to maximally deviate victim policies and identify the optimal attack strategy. (2) The \emph{cooperation} problem, which arises as merely perturbing victim actions arbitrarily or toward locally suboptimal cases is insufficient to indicate the failure of cooperative victims. The adversary faces the challenge of exploring and deceiving victims into a long-term, jointly detrimental failure scenario.

To address these challenges, we introduce \emph{Adversarial Minority Influence (AMI)}, a black-box policy-based attack for c-MARL. Shown in Fig. \ref{motivation}, AMI differs from other attacks by deceiving victims in an \emph{unilateral} and \emph{targeted} manner. The term \emph{minority influence} originates from social psychology \cite{crano2007minorityinfluence1}, wherein minorities (adversary) can \emph{unilaterally} sway majorities (victims) to adopt its own \emph{targeted} belief. Technically, to maximally deviate victim policies amidst complex agent interactions, we quantify and maximize the adversarial impact from the adversary to each victim. Our \emph{unilateral influence filter} adapts mutual information, a bilateral agent-wise relation metric, into a unilateral relation metric that represents the influence from the adversary to the victim. This is achieved by decomposing mutual information and filtering out the harmful influence from victims to the adversary. To deceive victims into a jointly worst-case failure, we learn a long-term, worst-case target for each victim, which leads to collective failure. The target is determined by the \emph{targeted adversarial oracle}, a reinforcement learning agent that generates worst-case target actions for each victim by co-adapting with our adversarial policy. Ultimately, the attacker modifies its policy to direct victims toward these adversarial target actions through unilateral influence. Our \textbf{contributions} are listed as follows:

\begin{itemize}


    \item We develop AMI, a strong and practical attack towards c-MARL, which leverages the intricate influence among agents and the cooperative nature of victims in c-MARL.
    

    \item We introduce the \emph{unilateral influence filter} and \emph{targeted adversarial oracle} to optimize the deviation of victim policies and deceive victims into suboptimal cooperation, thereby ensuring a potent attack capability.
    
    


    \item AMI achieves the first successful attack against real world robot swarms and effectively fool agents in simulation environments into collectively worst-case scenarios, including StarCraft II and Multi-agent Mujoco.

\end{itemize}

\section{Related Work}
\subsection{Overview of Adversarial Attacks}


Initially proposed in the field of computer vision, adversarial attacks consist of carefully crafted perturbations that, while imperceptible to humans, can deceive deep neural networks (DNNs) into making incorrect predictions \cite{szegedy2013fgsm, goodfellow2014fgsm, cwattack}. Given a DNN $F_{\theta}$, a clean image $\mathbf x$, a perturbed image $\mathbf x_{adv}$, and the ground truth label $y$, an adversarial example can be formulated as follows:
\begin{equation}
\mathbb F_{\theta} \left (\mathbf \mathbf x_{adv} \right ) \neq y  \quad s.t. \quad \|\mathbf x-\mathbf x_{adv}\| \leq \epsilon.
\end{equation}
In this formulation, $\| \cdot\|$ represents a distance metric used to constrain the distance between $\mathbf{x}$ and $\mathbf{x}_{adv}$ by $\epsilon$. Subsequently, it was demonstrated that reinforcement learning (RL) is also susceptible to adversarial attacks \cite{huang2017statebased1, kos2017statebased3, lin2017statebased2}. Owing to the sequential decision-making nature of RL, adversarial attacks in this context aim to generate a perturbation policy $\pi^\alpha$ that minimizes the victim's cumulative reward $\sum_t \gamma^t r_t$, which can be expressed as:
\begin{equation}
\min_{\pi^\alpha} \sum_t \gamma^t r_t.
\end{equation}
Adversarial attacks are important to distinguish as test-time attacks, where the adversary targets a specific victim without participating in the training process. As another line of research, training-time attacks interfere with victim training, resulting in trained victims either failing to perform well (poisoning attack) \cite{huang2019reward1, wu2022copa} or executing adversary-specified actions when specific triggers are present (backdoor attack) \cite{behzadan2019backdoor1, kiourti2019trojdrl, wang2021backdoor2}. Note that our method is a test-time attack, and is not related to training-time attacks.

\subsection{RL Attacks by Observation Perturbation}

Test-time perturbation of RL observations can deceive the policy of RL agents, causing them to execute suboptimal actions and fail to achieve their goals. For single-agent RL attacks, early research employed heuristics such as preventing victims from selecting the best action \cite{huang2017statebased1} or choosing actions with the lowest value at critical time steps \cite{kos2017statebased3, lin2017statebased2}. Later work framed the adversary and victim within an MDP \cite{zhang2020samdp}, enabling the optimal observation perturbation to be learned as an action within the current state using an RL agent \cite{zhang2021iclrATLA, sun2021optimaladv}. For c-MARL attacks, Lin et al. \cite{lin2020jieyulins&p} proposed a two-step attack that first learns a worst-case attack policy and then employs a gradient-based attack \cite{papernot2016jsma} to execute it. \cite{zan2023obsbased} generated attacks on one victim and transfer it to the rest of the victims. However, they assume attacker can modify victim observations and has white-box access to victim parameters, which can be impractical in real world Another line of research, termed adversarial communication, targets communicative c-MARL by sending messages to victim agents that cause failure upon receiving the adversarial message. Adversarial messages can be added to representations \cite{tu2021advcommunication1} or learned by the adversary \cite{xue2021misspokeormislead, blumenkamp2021advcommunication2, mitchell2020advcommunication3}. However, these methods are inapplicable when victim agents do not communicate, a common assumption in many mainstream c-MARL algorithms \cite{rashid2018qmix, yu2021mappo, lowe2017maddpg}.

\subsection{RL Attacks by Adversarial Policy}


Distinct from observation-based attacks, adversarial policy attacks do not necessitate access to victim observations or parameters (black-box). Rather, they introduce an adversarial agent whose actions are designed to deceive victim agents, causing them to take counterintuitive actions and ultimately fail to achieve their goals. In this paper, the terms ``policy-based attack" and ``adversarial policy" are used interchangeably. Gleave et al. \cite{gleave2019iclr2020advpolicy} were the first to introduce adversarial policy in two-agent zero-sum games. This approach was latter applied to multi-agent consensus game, where agents have similar, but non-identical objectives \cite{figura2021consensusacc}. Subsequent research has enhanced adversarial policy by exploiting victim agent vulnerabilities. Wu et al. \cite{wu2021usenix} induced larger deviations in victim actions by perturbing the most sensitive feature in victim observations, identified through a saliency-based approach. However, larger deviations in victim actions do not necessarily correspond to strategically worse policies. Guo et al. \cite{guo2021icml2021} extended adversarial policies to general-sum games by simultaneously maximizing the adversary's reward and minimizing the victim's rewards. Yet, none of these studies have considered adversarial policies in c-MARL settings. To better evaluate the performance of our attack in multi-agent adversarial policy scenario, we adapt some observation-based attack in MARL \cite{zan2023obsbased} as baselines.

\section{Problem Formulation}

We conceptualize adversarial attacks targeting c-MARL agents within the framework of a partially observable Stochastic game (POSG) \cite{hansen2004posg}, which can be characterized by a tuple: 
\begin{equation}
\mathcal G = \langle \mathcal N,\mathcal S, \mathcal O, O, \mathcal A, \mathcal T, R, \gamma \rangle.
\end{equation}
Specifically, $\mathcal N = \{\mathcal N^\alpha, \mathcal N^\nu\} = \{1, ..., N\}$ is the set containing $N$ agents. Throughout this paper, we use $\alpha$ for adversaries and $\nu$ for victims. $\mathcal S$ is the global state space, $\mathcal{O} = \times_{i \in \mathcal{N}} \mathcal{O}^i$ is the observation space, $O$ is the observation emission function, $\mathcal{A} = \times_{i \in \mathcal{N}}\mathcal{A}_i$ is the action space, $\mathcal{T}: \mathcal{S} \times \mathcal{A} \rightarrow \Delta(\mathcal{S})$ is the state transition probability. The reward $R^\alpha: \mathcal{S} \times \mathcal{A} \rightarrow \mathbb{R}$ is shared for all adversaries and $R^\nu: \mathcal{S} \times \mathcal{A} \rightarrow \mathbb{R}$ is shared for all victims. $\gamma \in [0, 1)$ is the discount factor.

At each timestep, each agent $i$ observes $o_{t, i} = O(s_t, i)$ and add it to history $h_{t, i} = [o_{0, i}, a_{0, i}, ... ,o_{t, i}]$ to alleviate partial observability \cite{oliehoek2016decpomdp}. For victims, we assume its joint policies $\mathbf{\pi}^\nu(\mathbf a^\nu|\mathbf h) = \prod_{i \in \mathcal N^\nu} \pi^\nu(a_i^\nu|h_i)$ are fixed during deployment \cite{hofer2021sim2real} and take joint actions according to joint policies $\mathbf a^\nu \sim \mathbf{\pi}^\nu(\cdot|\mathbf h)$. For adversary, it takes actions by an adversarial policy $a^\alpha \sim \pi^\alpha(\cdot|h_i)$. Next, the game environment proceeds to the next state via transition probability $\mathcal T(s_{t+1}|s_t, a^\alpha, \mathbf a^\nu)$ and receive the reward for adversary, $r_t^\alpha = R^\alpha(s_t, a^\alpha_t, \mathbf{a}^\nu_t)$. The objective of the adversary is to learn an adversarial policy $\pi^\alpha$ to maximize its expected discounted cumulative reward, \ie, $J(\pi^\alpha)=\mathbb{E}_{s, a^\alpha, \mathbf{a}^\nu} [\sum_t \gamma^t r^\alpha_t]$. Since defender policies are fixed, the attacker must solve a reinforcement learning problem, but can exploit the weakness in policies of other agents by maximally deviating victims into a jointly worst-case situation.

\textbf{Assumptions.} To keep our attack practical, we assume attacker cannot manipulate victim observations or choose which agent to control. Additionally, attacker does not have the models (\ie, architectures, weights, gradients) and rewards of victims. Following the centralized training, decentralized execution (CTDE) paradigm for c-MARL attack \cite{lin2020jieyulins&p, tu2021advcommunication1, blumenkamp2021advcommunication2, mitchell2020advcommunication3}, we posit that adversaries have access to the state and reward only during training. However, during actual attack deployment, they must rely solely on their local observation histories. To implement our AMI attack in physical environment, we adopt a Sim2Real paradigm \cite{hofer2021sim2real}, such that the adversary first trains its adversarial policy within a simulated environment, then freezes the policy's parameters and deploys the attack to real-world robots.

\textbf{Applications.} Our adversary can be launched as a malicious participator in a multi-agent distributed system \cite{wu2018resourceallocation1, zhang2009resourceallocation3}. Adversaries can also hijack an agent directly \cite{giray2013hijacking, ly2021hijacking, rodday2016mitm} and utilize the controlled agent to take an adversarial policy. Apart from malicious use, our AMI attack also functions as a testing algorithm to evaluate the worst-case robustness of c-MARL algorithms, which helps managing algorithmic risks before deploying it in risk-sensitive applications.

\begin{figure*}[t]
\centering
\includegraphics[scale=0.45]{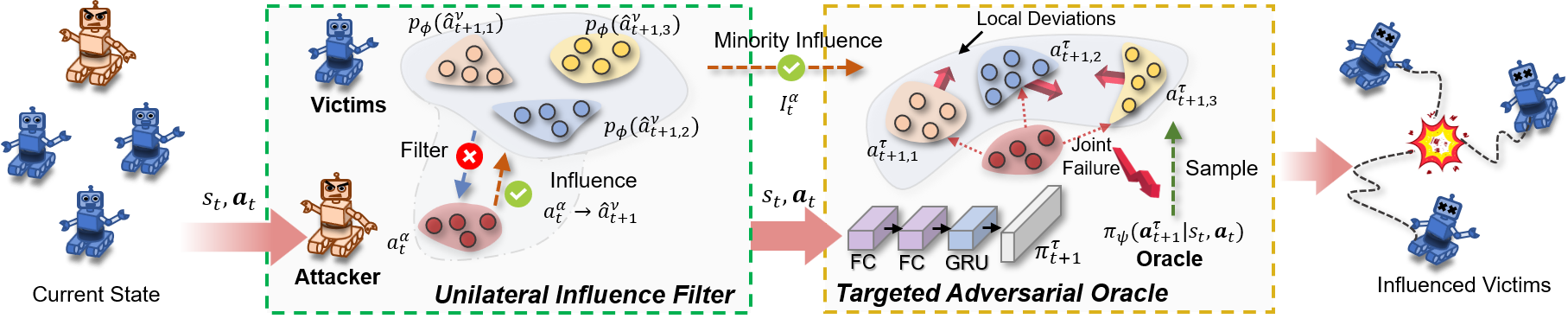}
\caption{Framework of AMI. Unilateral influence filter decompose mutual information into minority influence and majority influence terms, while keeping latter for asymmetric influence. Targeted adversarial oracle is an RL agent that generates a worst-case target for each victim. Attacking victims towards this target results in jointly worst cooperation.}
\label{framework}
\vspace{-0.2in}
\end{figure*}

\section{Method}

As previously mentioned, adversarial policy towards c-MARL presents two challenges: the \emph{influence} challenge, which necessitates the adversary to maximize victim policy deviations under intricate agent-wise interactions; and the \emph{cooperation} challenge, which requires the adversary to deceive agents into jointly worst failures. In this paper, we address these challenges through our \emph{adversarial minority influence (AMI)} framework. As depicted in Fig. \ref{framework}, AMI achieves potent attack capabilities by \emph{unilaterally} guiding victims towards a \emph{targeted} worst-case scenario. To maximize policy deviation of victims, we propose the \emph{unilateral influence filter} that characterizes the adversarial impact from the adversary to victims by decomposing the bilateral mutual information metric and eliminating the detrimental influence from victims to the adversary. To steer victims towards jointly worst actions, the \emph{targeted adversarial oracle} is a reinforcement learning agent that co-adapts with the adversary and generates cooperatively worst-case target actions for each victim at every timestep.

\subsection{Unilateral Influence Filter}


In a c-MARL attack, due to the intercorrelated nature of agent policies, targeting one victim inevitably impacts the policies of other victims as well. In light of such intricate relationships, maximizing policy deviations for victims necessitates that the adversary first characterizes the influence of its actions on each victim before maximizing that influence. Consequently, to delineate the unilateral influence from the adversary to the victim, we draw inspiration from the minority influence theory in social psychology \cite{crano2007minorityinfluence1}. 

Using mutual information as a starting point, we emphasize that the policies of the adversary and the victims are interdependent. However, mutual information fails to fully reflect the deviation in victim policy as it includes the reverse influence from the victims to the adversary, which is counterproductive for the attack. By focusing on the one-way influence from adversary to victims only, our approach aligns with the principles of minority influence, where a small, focused group can effectively influence the behavior of a larger group.



We begin by examining mutual information, a commonly employed bilateral influence metric in the c-MARL literature \cite{jaques2019icmlsocialinfluence, wang20192020iclrinfluence, fayad2021influencenips, li2022pmicinfluenceicml}, which captures the relationship between agents. For instance, considering social influence \cite{jaques2019icmlsocialinfluence}, the mutual information between the adversary action $a_t^\alpha$ and the influenced victim action $a_{t+1, i}^\nu$ can be expressed as $I(a_t^\alpha; a_{t+1, i}^\nu | s_t, \mathbf{a}_t^\nu)$, where $a_t^\alpha$ denotes the action taken by adversary $\alpha$ at time $t$, and $a_{t+1, i}^\nu$ represents the action taken by the $i^{th}$ victim $\nu$ at time $t+1$. Since victim observation and parameter is unknown, the action probability of $a_{t+1, i}^\nu$ can be approximated by network $p_\phi$ in a supervised learning objective, \ie, $\max_{\phi} \sum_{i=1}^{\mathcal N^\nu} \sum_{t=0}^{T-1} \log p_\phi(\hat{a}_{t+1, i}^\nu|s_t, \mathbf{a}_{t})$,  where $T$ denotes the total timesteps of an episode. In the remainder of our paper, we use $\hat{a}$ to signify that the action is predicted, rather than ground truth.

Although mutual information is capable of characterizing agent-wise influence in c-MARL, it is important to note that \emph{attacks towards c-MARL differ from cooperation due to the fixed nature of victim parameters}, which renders victim actions more challenging to modify compared to those of the adversary. To demonstrate the effects of fixed victim policy, we decompose the mutual information between the adversary action $a_t^\alpha$ and a victim agent $a_{t+1, i}^\nu$ as follows:
\begin{equation}
\label{eqn:mutualinfo}
\begin{split}
I(a_t^\alpha; \hat{a}_{t+1, i}^\nu | s_t, \mathbf{a}_t^\nu) = & \underbrace{-H(\hat{a}_{t+1, i}^\nu|s_t, a_t^\alpha, \mathbf{a}_t^\nu)}_{\text{majority}}\\
& + \underbrace{H(\hat{a}_{t+1, i}^\nu|s_t, \mathbf{a}_t^\nu)}_{\text{minority}}.
\end{split}
\end{equation}
In the context of a c-MARL attack, we refer to the first term as \emph{majority influence}, that is, the extent to which the minority (attacker) adapts its policy to conform with the policies of the majority (victims). In order to maximize mutual information, $H(\hat{a}_{t+1, i}^\nu|s_t, a_t^\alpha, \mathbf{a}_t^\nu)$ should be \emph{minimized}, such that having knowledge of $a_t^\alpha$ reduces the uncertainty surrounding the victim policy. In policy-based attacks within c-MARL, the majority influence term in mutual information causes attackers to comply with victim policies, yielding high mutual information but weak attack capability. To elucidate this, consider that the parameters of victims are fixed, whereas the adversary policy is learned. We assume that it is significantly more straightforward to modify the adversary's policy than the victim's policy. Consequently, in order to minimize majority influence, the most effective approach for an attacker is to adjust its action $a_t^\alpha$ to render it more predictive of victim actions, without actually altering it.





Simultaneously, the second term, $H(\hat{a}_{t+1, i}^\nu|s_t, \mathbf{a}_t^\nu)$, can be interpreted as a form of \emph{minority influence} devoid of victim impact, reflecting solely the adversary's effect on victims. This term accounts for the entropy of the victim policy without conditioning on $a_t^\alpha$, thereby establishing an \emph{unilateral} metric. Given that the influence of the adversary action $a_t^\alpha$ is marginalized, the adversary is unable to modify its action to cater to the policy of victims.

Building upon the aforementioned discourse, we examine and generalize minority influence within mutual information to enhance attack capabilities. In particular, maximizing minority influence requires current adversary policy to have large uncertainty in victim policy:
\begin{equation}
\begin{split}
\label{eqn:klrandom}
H(\hat{a}_{t+1, i}^\nu|s_t, \mathbf{a}_t^\nu) &=  - D_{KL}\left(p_\phi(\hat{a}_{t+1, i}^\nu|s_t, \mathbf{a}_t^\nu)|| \ \mathcal{U})\right) + c \\
&= - D_{KL}\left(\mathbb{E}_{\tilde{a}_t^\alpha \sim \pi_t^\alpha}\left[p_\phi(\hat{a}_{t+1, i}^\nu|s_t, \tilde{a}_t^\alpha, \mathbf{a}_t^\nu)\right]|| \ \mathcal{U})\right) + c,
\end{split}
\end{equation}
see Appendix. \ref{appendix_proofa} for detailed derivation. In the equation, $\mathcal{U}$ represents a uniform distribution for victim actions applicable to both discrete and continuous action spaces, $D_{KL}$ is the Kullback-Leibler divergence, $c$ is a constant, $\tilde{a}_t^\alpha$ refers to the counterfactual action sampled from the adversarial policy $\pi^\alpha$.


Equation \ref{eqn:klrandom} shows the second term is equivalent to minimizing the KL divergence between the victim policy and the uniform distribution. For a more general relation metric between adversary and victim, we can release the constraints of Eqn. \ref{eqn:klrandom} and replace $\mathcal{U}$ by a worst-case target distribution $\mathcal{D}$ and generalize $D_{KL}$ to any distance metric $d(\cdot, \cdot)$. Consequently, the unilateral influence can be expressed as:
\begin{equation}
\label{eqn:unilateral}
I = d\left(\mathbb{E}_{\tilde{a}_t^\alpha \sim \pi^\alpha}\left[p_\phi(\hat{a}_{t+1, i}^\nu|s_t, \tilde{a}_t^\alpha, \mathbf{a}_t^\nu)\right], \ \mathcal{D})\right).
\end{equation}
In this manner, influencing victims unilaterally is tantamount to minimizing the distance between the expected victim policy under the adversary policy and a target distribution.


\subsection{Targeted Adversarial Oracle}


Upon elucidating the maximization of victim policy deviations through unilateral influence, it remains crucial to ensure that victims are guided toward an optimal target. Merely deviating the victim policy arbitrarily or in the direction of a locally inferior case does not guarantee a globally worst-case failure for c-MARL. To enhance attack capability, it is necessary to ascertain globally worst target actions for victims and subsequently steer each victim towards its target $\mathcal{D}$ (Eqn. \ref{eqn:unilateral}) using the proposed unilateral influence approach. Consequently, we introduce a reinforcement learning agent that learns these jointly worst target actions by co-adapting its policy with the attacker in a trial-and-error process.

To accomplish the targeted attack objective, we introduce the \emph{Targeted Adversarial Oracle (TAO)}, a reinforcement learning agent $\pi^\tau$ that guides the attacker to influence each victim toward their globally worst-case direction. To achieve this, TAO use global state as input, and is used to guide the attacker \emph{only} in training. During execution, the attacker acts on its own without the guidance of TAO. As an RL agent, TAO adjusts to the current perturbation budget of the adversary through a trial-and-error process: if the adversary can significantly impact victim policies, TAO can generate the target more aggressively, directing victims to undertake riskier actions, thus achieving greater attack capability; conversely, if the adversary has limited influence on victims, TAO strives to introduce minor yet effective perturbations to victims, which is feasible under the current perturbation budget. Analogous to the attacker, TAO's objective is to maximize the adversary's goal. We define the following Bellman operator $\mathcal{B}^{\tau}$ to update the value function of TAO:
\begin{equation}
\label{eqn:tao_bellman_operator}
\begin{split}
\left(\mathcal{B}^{\tau} Q^{\tau}\right)(s, a^{\tau}, a^\alpha) & =  r_t^\alpha + \gamma \sum_{s'} \mathcal{T}(s'|s, a^\alpha, \mathbf{a}^\nu) \sum_{a'^\tau \in \mathcal A} \\
& \pi(a'^\tau|s') \hspace{-18pt}\sum_{\hspace{18pt}(\mathbf{a}'^\nu, a'^\alpha) \in \mathcal A} \hspace{-18pt} \pi^\alpha(a'^\alpha|h_i') \pi^\nu(\mathbf{a}'^\alpha|\mathbf{h}') Q^\tau(s', a'^{\tau}, a'^\alpha).
\end{split}
\end{equation}
To avoid non-stationarity, $Q^{\tau}$ also depends on $a^\alpha$, the adversary's action. While $a^{\tau}$ does not directly interact with the environment, the attacker's policy is \emph{implicitly} influenced by $a^{\tau}$ through the unilateral influence term added to attacker's reward. Assuming the space of state, action of TAO and adversary are finite, we can proof $\mathcal{B}^{\tau}$ is a contraction operator. Consequently, updating $Q^{\tau}(s, a^{\tau}, a^\alpha)$ by $\mathcal{B}^{\tau}$ will converge to optimal value $Q^{\tau, *}(s, a^{\tau}, a^\alpha)$. See the proof in Appendix. \ref{appendix_proofb}.

In practice, we use proximal policy optimization (PPO) \cite{schulman2017ppo} to optimize the policy of TAO. The value function used by PPO is an advantage function $A_t^\tau$, with the policy of TAO $\pi^{\tau}$ updated by PPO objective:
\begin{equation}
\label{eqn:ppo_adversarial_oracle}
\begin{split}
&\max_{\pi^{\tau}} \mathbb{E}_{\pi^{\tau}}\left[\text{min}\left(\text{clip}\left(\rho_t, 1-\epsilon, 1+\epsilon\right){A}_{t}^{\tau}, \rho_t {A}_{t}^{\tau}\right) \right], \\
& \text{where} \ \ \rho_t = \frac{1}{N^\nu}\sum_{i=1}^{|\mathcal N^\nu|}\frac{\pi^{\tau}(a_{t+1, i}^\tau|s_t, \mathbf{a}_t^\nu, a_t^\alpha)}{\pi^\tau_{old} (a_{t+1, i}^\tau|s_t, \mathbf{a}_t^\nu, a_t^\alpha)},
\end{split}
\end{equation}
where $\pi_{old}^{\tau}$ and $\pi^{\tau}$ represent the previous and updated policies for TAO, respectively. ${A}_{t}^{\tau}$ is the advantage function determined by the generalized advantage estimation (GAE) \cite{schulman2015GAE}. The function $\textrm{clip}(\rho_t, 1-\epsilon, 1+\epsilon)$ constrains the input $\rho_t$ within the limits of $1-\epsilon$ and $1+\epsilon$. $|\mathcal N^\nu|$ is the number of victims. In this case, the policy $\pi^{\tau}(a_{t+1, i}^\tau|s_t, \mathbf{a}_t^\nu, a_t^\alpha)$, as determined by TAO, functions as the desired target distribution $\mathcal{D}$ in Eqn. \ref{eqn:unilateral} for victim $i$ at time $t$. Notably, as PPO is an on-policy algorithm, we derive $\mathcal D$ by sampling a target action $a_{t+1, i}^{\tau}$ from $\pi^{\tau}$ instead.


\subsection{Overall Training}


In summary, the influence metric $I_t^\alpha$ used by AMI at time $t$ combines the optimal target action $a_{t+1, i}^{\tau}$ generated by TAO as the target distribution $\mathcal{D}$ for unilateral influence in Eqn. \ref{eqn:unilateral}:
\begin{equation}
\label{eqn:advinf_final}
\begin{split}
I_t^\alpha = \sum_{i=1}^{N^\nu}&\biggl[d\Bigl(\mathbb{E}_{\tilde{a}_t^\alpha \sim \pi^\alpha} \Bigl[p_{\phi}(\hat{a}_{t+1, i}^\nu|s_t, \tilde{a}_t^\alpha, \mathbf{a}_t^\nu)\Bigr], a^{\tau}_{t+1, i} \sim \pi^{\tau}(\cdot|s_t, a_t^\alpha, \mathbf{a}_t^\nu)\Bigr)\biggr],
\end{split}
\end{equation}
In the case of $d(\cdot, \cdot)$ as the distance function for AMI, the calculation differs depending on the control setting. For \emph{discrete control}, the distance function is computed as $d(p(\hat{a}_{i}^{\nu}|s, \mathbf{a}), a_{i}^\tau) = -|| p(\hat{a}_{i}^\nu|s, \mathbf{a}) - \mathbf{1}_{\mathcal{A}}(a_{i}^\tau)||_{1}$, where $p(\hat{a}_{i}^\nu|s, \mathbf{a})$ is a shorthand for $\mathbb{E}_{\tilde{a_t^\alpha} \sim \pi^\alpha}\left[p_{\phi}(\hat{a}_{t+1, i}^\nu|s_t, \tilde{a}_t^\alpha, \mathbf{a}_t^\nu)\right]$ and $a_{i}^\tau$ denotes $a_{t+1, i}^\tau$. $\mathbf{1}_{\mathcal{A}}(a_{i}^\tau) = [a_{i}^\tau \in \mathcal{A}^\tau_i]$ is the indicator function with action space $\mathcal{A}_i^\tau$. A negative sign is added to minimize the distance between the one-hot target action $\mathbf{1}_{\mathcal{A}}(a_{i}^\tau)$ and the victim action probability. In the case of \emph{continuous control}, the distance function is calculated as $d(p(\hat{a}_{i}^\nu|s, \mathbf{a}), a_{i}^\tau) = p(a_{i}^\tau|s, \mathbf{a})$, such that the target action $a_{t+1, i}^\tau$ has a high probability in the estimated victim action probability distribution. As verified in ablations, while we tune the distance functions for best performance, our method is not sensitive to the choice of distance functions.


In the final step, $I_t^\alpha$ serves as an auxiliary reward that is optimized by the policy of the adversarial agent $\pi^\alpha(a_t^\alpha|h^\alpha_t)$. The reward $r^{AMI}_t$ for the adversarial agent $\pi_\theta$ to optimize can be expressed as:
\begin{equation}
\label{eqn:reward_calculation}
r^{AMI}_t = r^\alpha_t + \lambda \cdot I^\alpha_t,
\end{equation}
In this case, $\lambda$ is a hyperparameter that balances the trade-off between the adversary reward $r^\alpha_t$ and maximizing the influence $I^\alpha_t$ on the victim agents. We define the following Bellman operator $\mathcal B^\alpha$ to update the value function of adversary's policy:
\begin{equation}
\label{eqn:adv_bellman_operator}
\begin{split}
\left(\mathcal{B}^\alpha Q^{\alpha}\right)(s, a^{\tau}, a^\alpha)&  =  r^\alpha + \lambda \cdot I^\alpha+ \gamma \sum_{s'} \mathcal{T}(s'|s, a^\alpha, \mathbf{a}^\nu) \\
& \sum_{a'^\tau \in \mathcal A} \pi(a'^\tau|s') \hspace{-18pt}\sum_{\hspace{18pt}(\mathbf{a}'^\nu, a'^\alpha) \in \mathcal A} \hspace{-18pt} \pi^\alpha(a'^\alpha|h_i') \pi^\nu(\mathbf{a}'^\alpha|\mathbf{h}') Q^\alpha(s', a'^{\tau}, a'^\alpha).
\end{split}
\end{equation}
Here, the dependency on $a^\tau$ is added to avoid non-stationarity. Using similar techniques for proofing the convergence of $\mathcal B^\tau$, we can proof $\mathcal B^\alpha$ is a contraction operator, thus updating $Q^{\alpha}(s, a^{\tau}, a^\alpha)$ via $\mathcal B^\alpha$ converge to the optimal value $Q^{\alpha, *}(s, a^{\tau}, a^\alpha)$. See the proof in Appendix. \ref{appendix_proofc}.

In practice, we compute the advantage function ${A}_{t}^{\alpha}$ using $r^{AMI}_t$ via GAE \cite{schulman2015GAE} and train the adversary using PPO \cite{schulman2017ppo}:
\begin{equation}
\label{eqn:ppo_adv_agent}
\begin{split}
& \max_{\pi^{\alpha}}  \mathbb{E}_{\pi^{\alpha}}\left[\min\left(\text{clip}\left(\rho_t, 1-\epsilon, 1+\epsilon\right){A}_{t}^{\alpha}, \rho_t {A}_{t}^{\alpha}\right) \right], \\
& \text{where} \ \ \rho_t = \frac{\pi^{\alpha}(a_{t}^\alpha|h_t^\alpha)}{\pi_{old}^{\alpha}(a_{t}^\alpha|h_t^\alpha)}.
\end{split}
\end{equation}
The complete training procedure is outlined in Algorithm \ref{alg1}.

\begin{algorithm}[t]
\caption{Adversarial Minority Influence Algorithm.}
\label{alg1}
\begin{algorithmic}[1]
\renewcommand{\algorithmicrequire}{\textbf{Input:}}
\renewcommand{\algorithmicensure}{\textbf{Output:}}
\REQUIRE Policy of victims $\pi^\nu$, adversary $\pi^\alpha$ and targeted adversarial oracle (TAO) $\pi^{\tau}$. Value function of adversary $V^\alpha$ and TAO $V^{\tau}$. Opponent modelling network $p_\phi$.
\ENSURE Trained policy network of adversary agent $\pi_\theta$.
\FOR{k = 0, 1, 2, ... K}
\STATE Perform rollout using current adversarial policy network $\pi^\alpha$ and TAO agent $\pi^{\tau}$. Collect a set of trajetories $\mathcal{D}_k = {\tau_i}$, where i = 1, 2, ..., $|\mathcal{D}_k|$.
\STATE Update opponent modelling model $p_\phi$.
\STATE Calculate advantage function of TAO ${A}_{t}^{\tau}$ by GAE, using value function $V^\alpha$ and adversary reward $r^{\alpha}$; Update value function network $V^{\tau}$ of TAO.
\STATE Update the policy network $\pi^{\tau}$ of TAO using Eqn. \ref{eqn:ppo_adversarial_oracle};
\STATE Calculate reward $r^{AMI}_t$ for adversary by Eqn. \ref{eqn:reward_calculation}.
\STATE Calculate advantage function ${A}_{t}^{\alpha}$ by GAE, using value function $V^{\alpha}$ and reward $r^{AMI}$; Update value function network $V^{\alpha}$ of adversary.
\STATE Update policy network $\pi^\alpha$ of adversary by Eqn. \ref{eqn:ppo_adv_agent}.
\ENDFOR
\end{algorithmic}
\end{algorithm}
\vspace{-0.1in}
\section{Experiments}


In this section, we perform comprehensive experiments in both simulated and real-world environments to assess the effectiveness of our AMI approach in terms of attack capability.

\subsection{Experimental Setup}

\subsubsection{Environments}


We assess the effectiveness of AMI in three distinct environments: (1) A real-world multi-robot rendezvous environment, in which robot swarms learn to gather together, known as \emph{rendezvous} \cite{huttenrauch2019swarm}. (2) StarCraft Multi-Agent Challenge (SMAC) \cite{samvelyan2019smac}, involving discrete control across six tasks, where the objective is to control a group of agents in a StarCraft game to defeat an opposing group; (3) Multi-Agent Mujoco (MAMujoco) \cite{peng2021facmac} for continuous control, comprising six tasks that require controlling robotic joints to optimize speed in a specific direction. All victim policies were trained using MAPPO \cite{yu2021mappo}. During the attack, the first agent is selected as the adversary.

\begin{figure}[!b]
    \centering
    \subfloat[E-puck2 robot]{
        \includegraphics[width=0.35\linewidth]{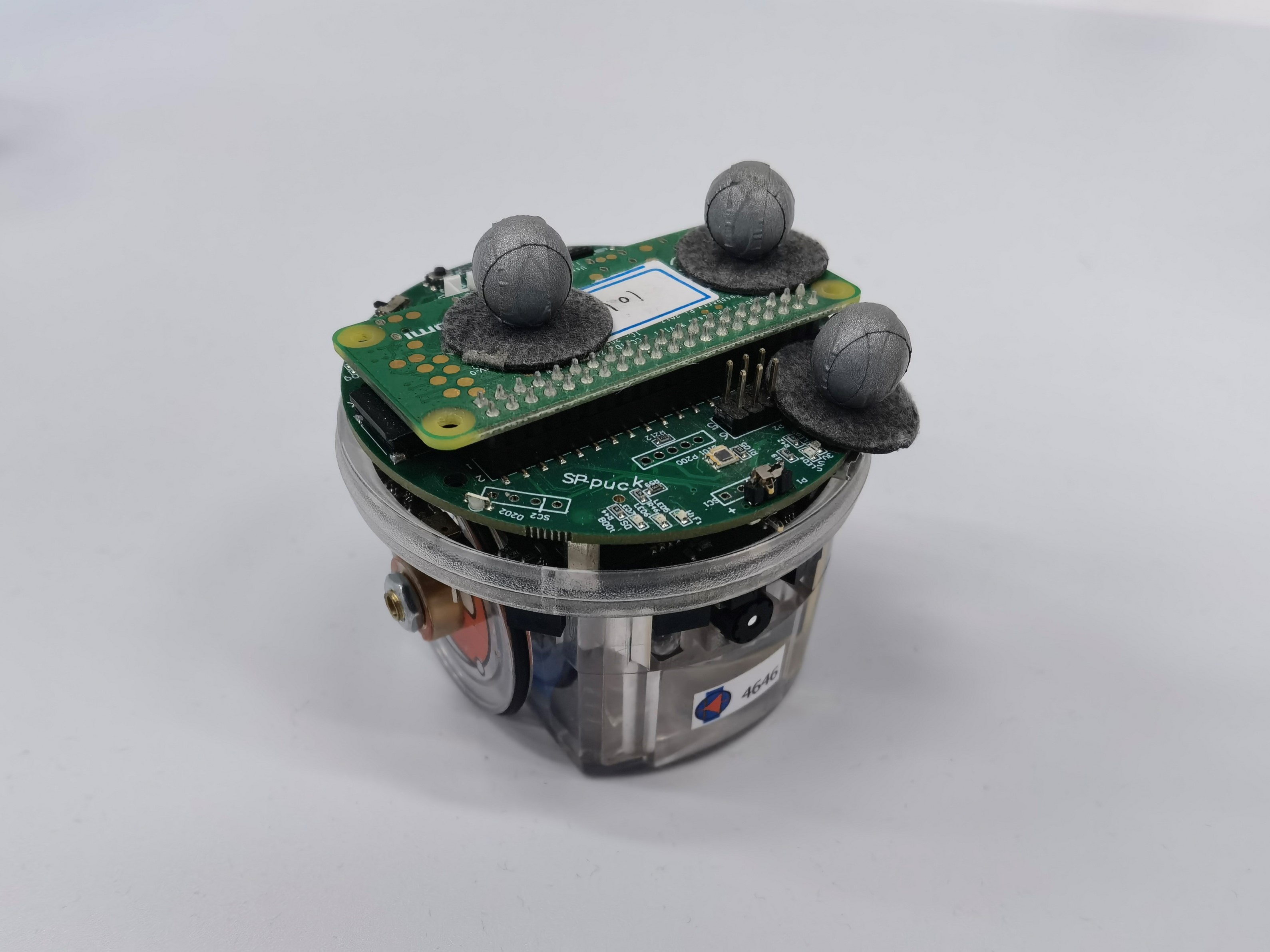}
        \label{fig:epuck}
    }
    \subfloat[Playground]{
        \includegraphics[width=0.35\linewidth]{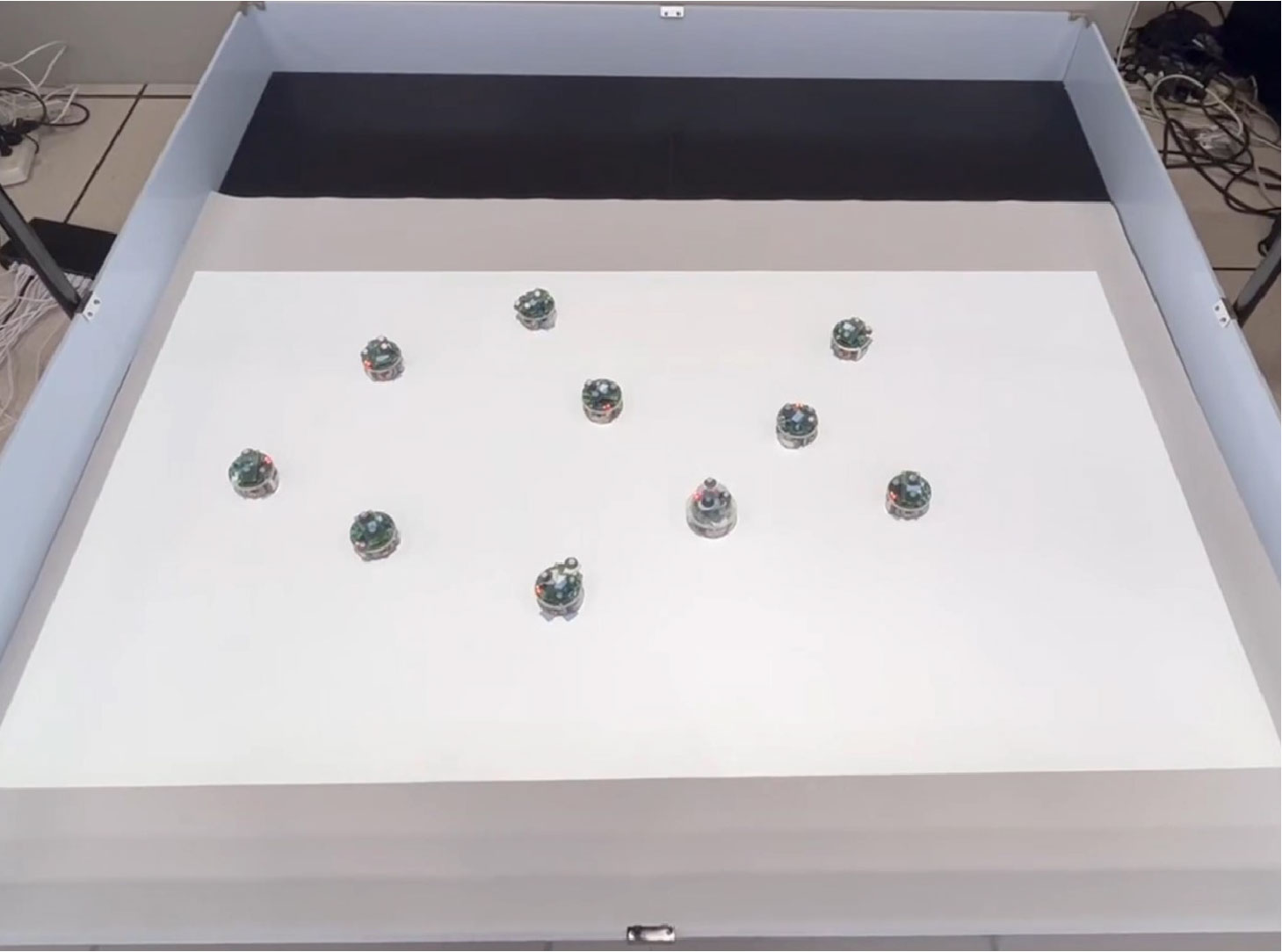}
        \label{fig:playground}
    }
    \caption{Illustration of the robot and playground for our real-world multi-robot rendezvous environment.}
    \vspace{-0.1in}
    \label{phys-env}
\end{figure}

\begin{figure}[!b]
    \centering
    \subfloat[Simulation Results]{
        \includegraphics[width=0.35\linewidth]{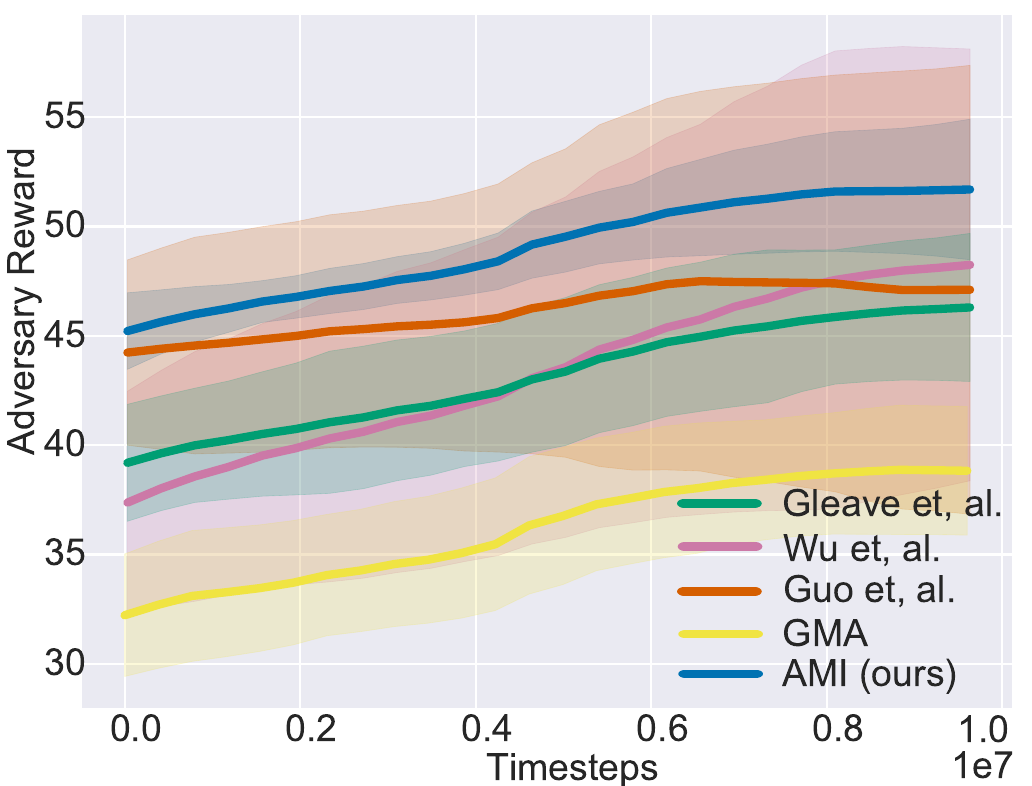}
        \label{subfig:robot}
    }
    \subfloat[Real World Results]{
        \includegraphics[width=0.35\linewidth]{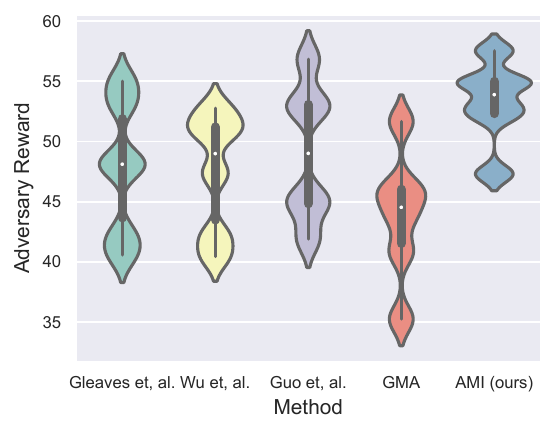}
        \label{subfig:robot-env}
    }
    \caption{Comparisons of AMI against baselines in simulation and real-world experiments.}
    \vspace{-0.1in}
    \label{phys-result}
\end{figure}

\begin{figure*}[!t]
    \centering
    \subfloat[\emph{Gleaves et al.}]{
        \includegraphics[width=0.18\linewidth]{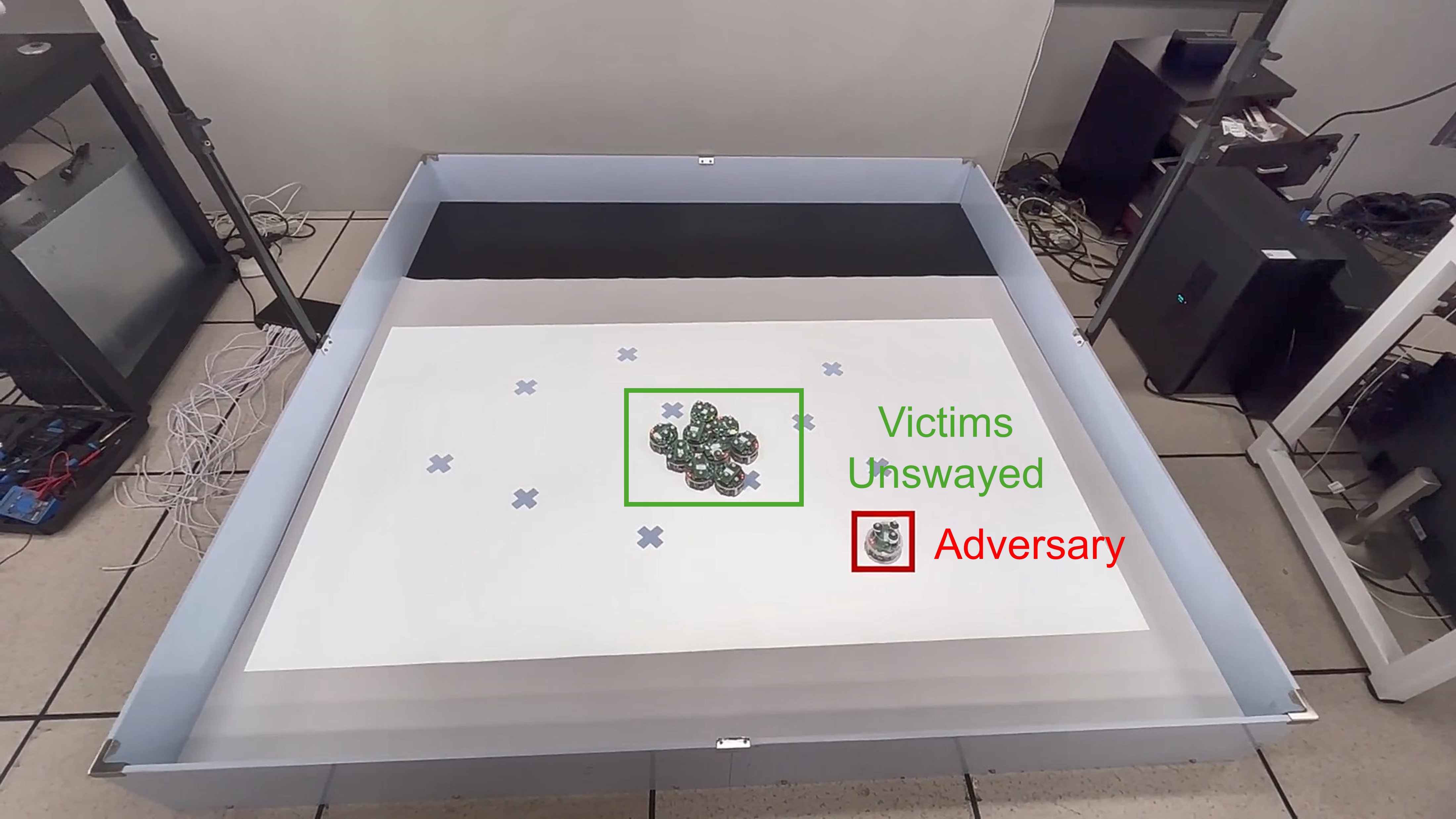}
        \label{subfig:phys-img-1}
    }
    \subfloat[\emph{Wu et al.}]{
        \includegraphics[width=0.18\linewidth]{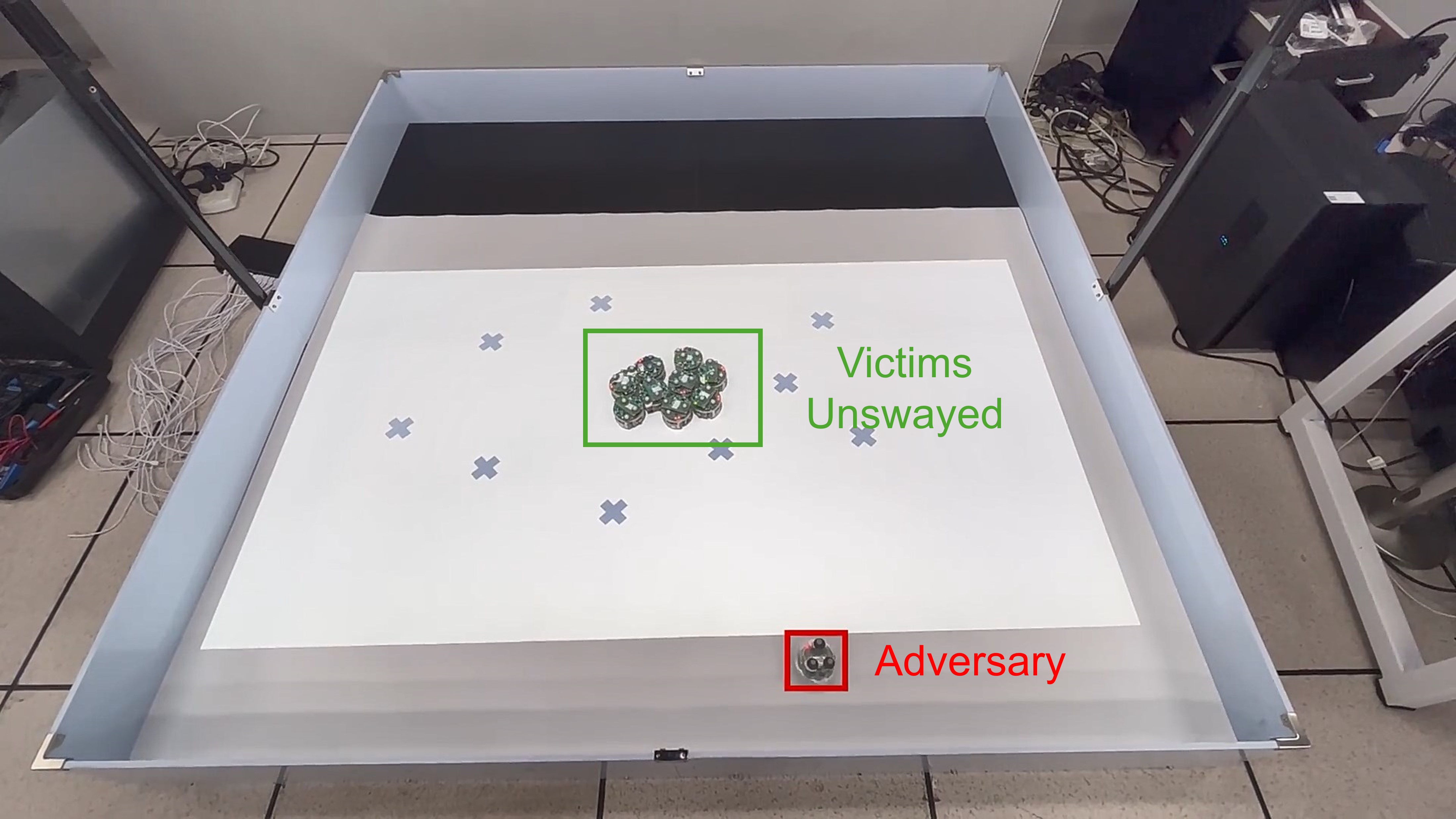}
        \label{subfig:phys-img-2}
    }
    \subfloat[\emph{Guo et al.}]{
        \includegraphics[width=0.18\linewidth]{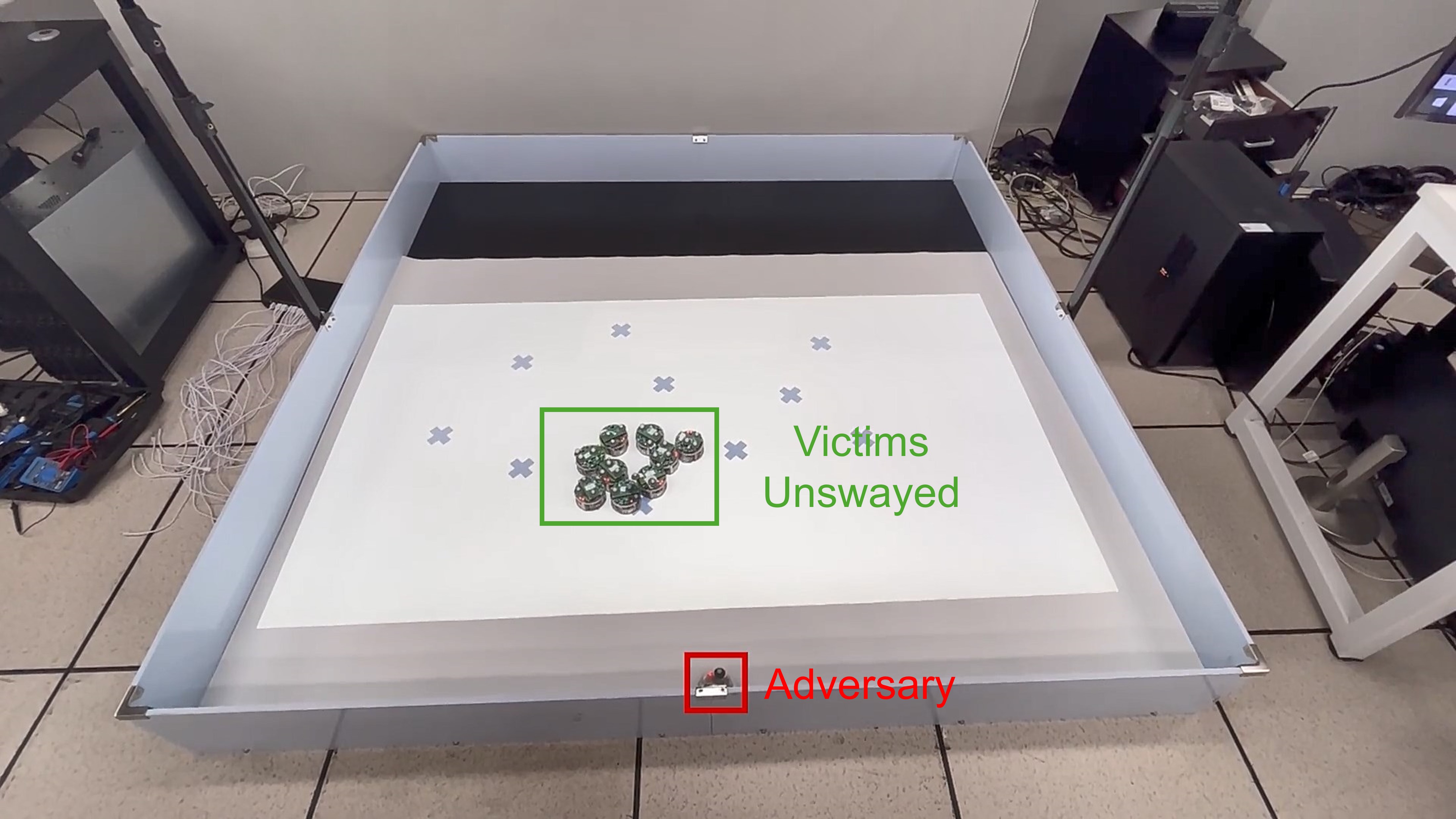}
        \label{subfig:phys-img-3}
    }
    \subfloat[\emph{GMA}]{
        \includegraphics[width=0.18\linewidth]{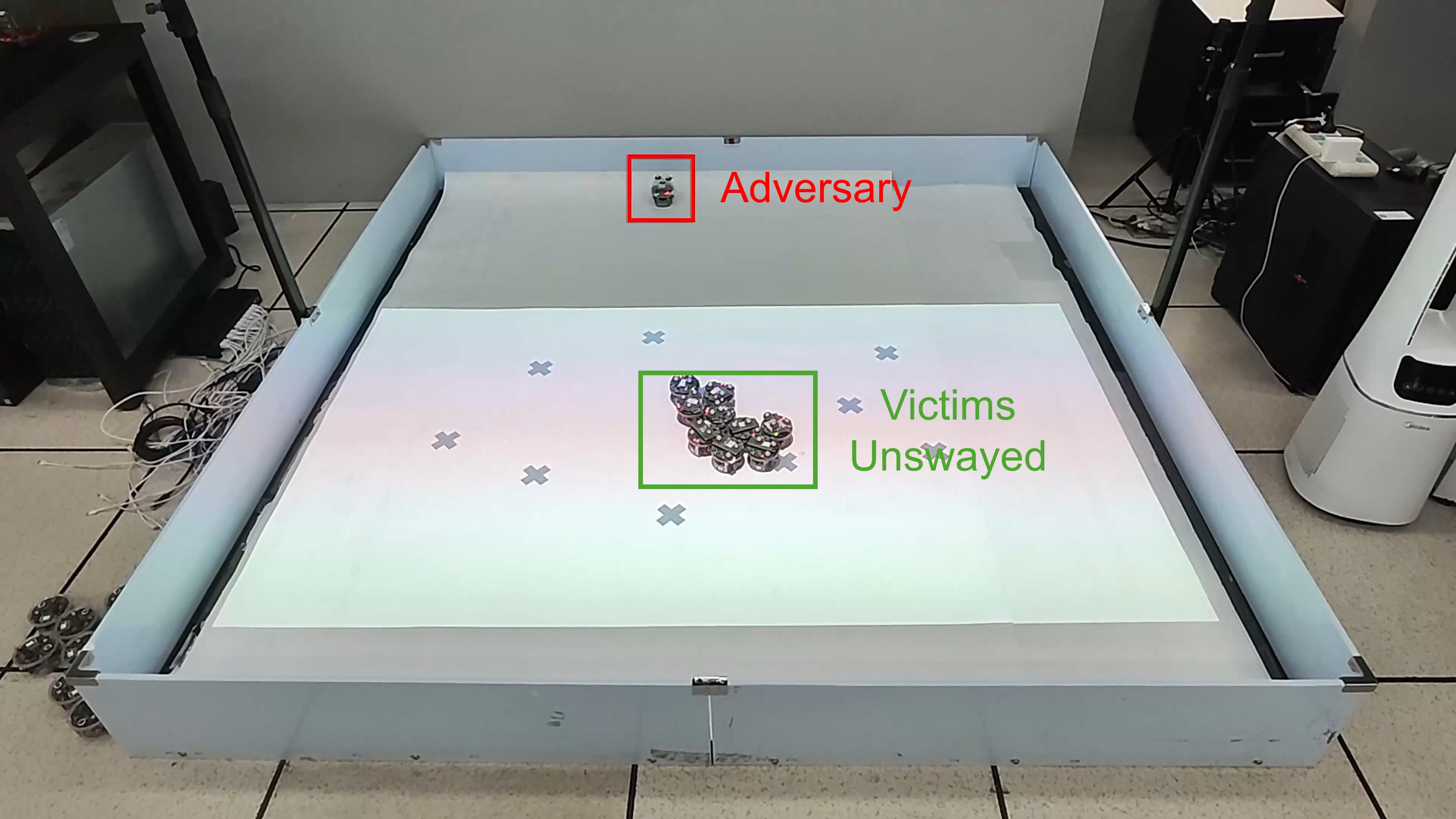}
        \label{subfig:phys-img-3}
    }
    \subfloat[AMI (ours)]{
        \includegraphics[width=0.18\linewidth]{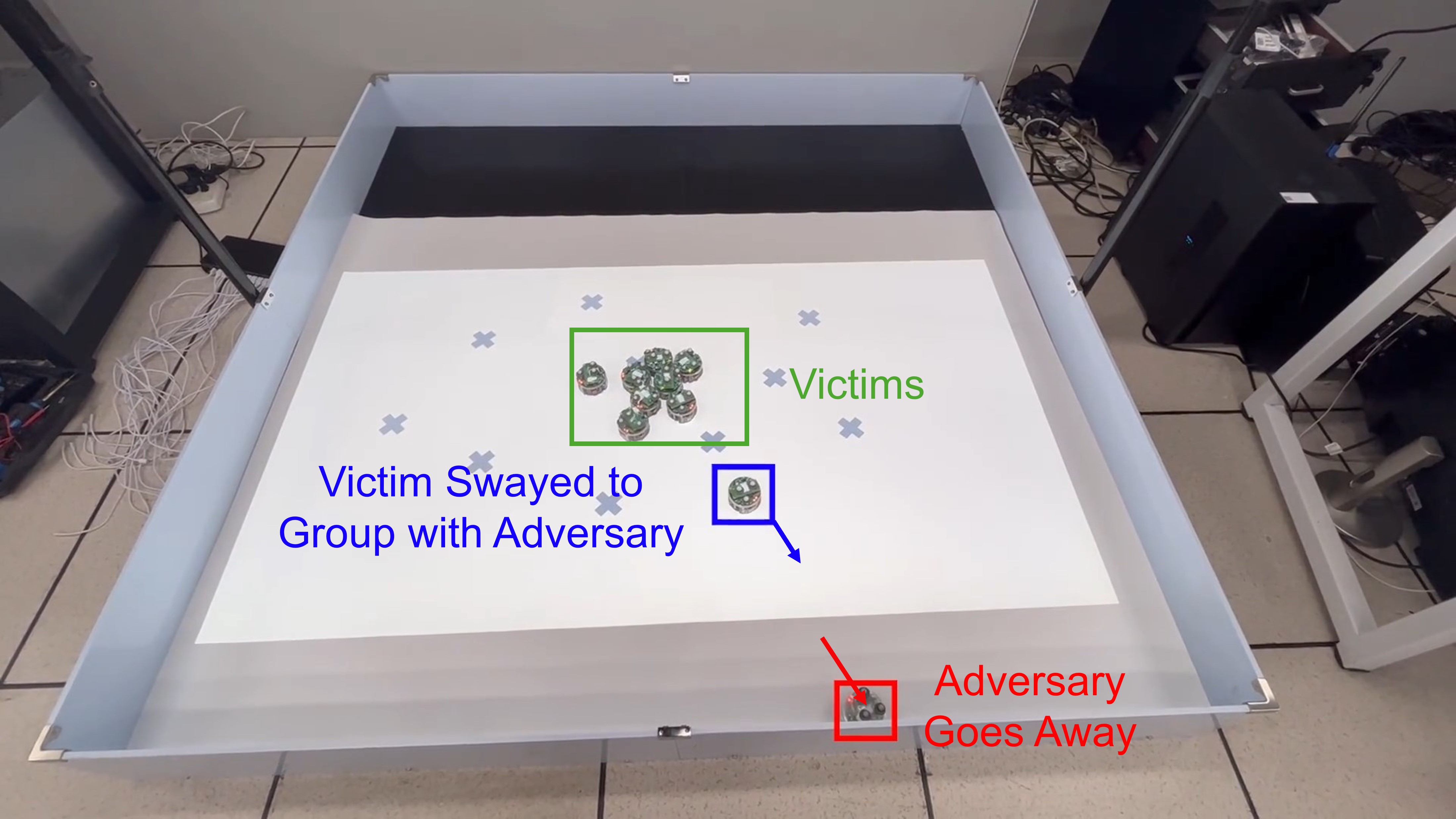}
        \label{subfig:phys-img-4}
    }
    \caption{Behaviors of robot swarms under our AMI attack, adversary indicated by red square. Our adversary is the only one to fool away an agent to group with our adversary.}
    \vspace{-0.1in}
    \label{phys-video}
\end{figure*}

\subsubsection{Compared methods and evaluation metrics}


We benchmark AMI against state-of-the-art adversarial policy methods, including single-agent adversarial policy, \emph{Gleave et al.} \cite{gleave2019iclr2020advpolicy}, \emph{Wu et al.} \cite{wu2021usenix}, \emph{Guo et al.} \cite{guo2021icml2021} and multi-agent attack GMA \cite{zan2023obsbased}. We adapt adversarial policy for single-agent case to multi-agent by substituting a single RL victim with multiple RL victims. As for multi-agent observation-based GMA attack, we allow their attack to manipulate victim action arbitrarily, same as our setting. To ensure a fair comparison, AMI and all baselines employ the same codebase, network structure, and hyperparameters. For method-specific hyperparameters, we tune their values for optimal performance. The adversary's goal is to maximize the adversarial reward $r^\alpha$, defined as (1) maximizing the loss of allies and minimizing the loss of enemies for SMAC. (2) minimizing the speed of agents in +x direction for MAMujoco. (3) maximizing the euclidean distance between all agents for rendezvous. All experiments were conducted with five random seeds, and results are presented with a 95\% confidence interval. The hyperparameters used for all experiments are listed in Appendix. \ref{appendix_hyperparam}.

\subsection{AMI Attack in Real World}

\begin{figure*}[!t]
    \centering
    \includegraphics[width=0.97\linewidth]{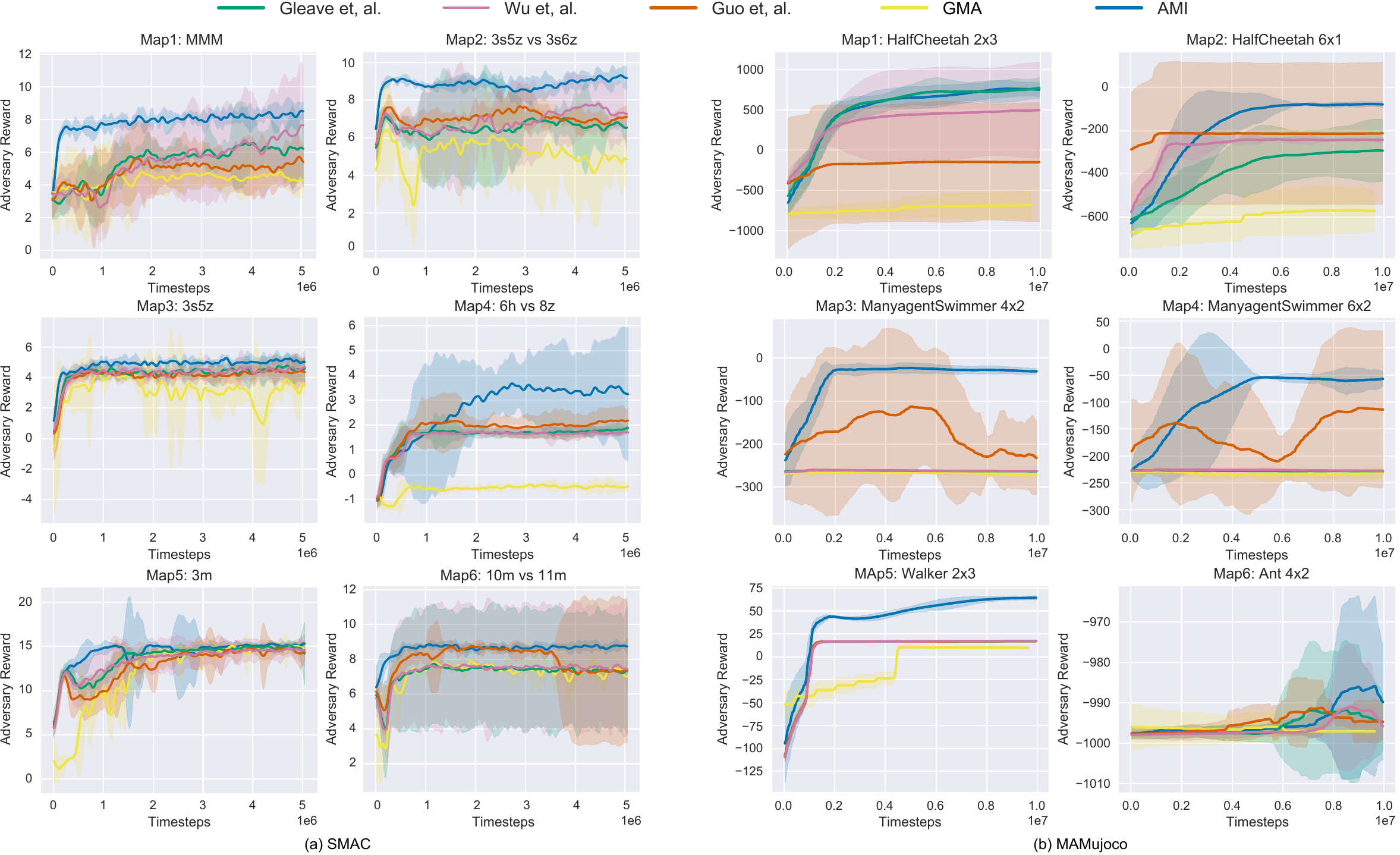}
        \label{fig:material}
    \caption{Learning curves for AMI and baseline attack methods in six SMAC and six MAMujoco environments. The solid curves correspond to the mean, and the shaped region represents the 95\% confidence interval over 5 random seeds.}
    \vspace{-0.1in}
    \label{exp_all}
\end{figure*}

To highlight the effectiveness of AMI, we evaluate AMI in real-world multi-robot environments. To the best of our knowledge, this is the first evaluation of adversarial policy in real world. As shown in Fig. \ref{phys-env}, we create an environment with 10 e-puck2 robots (Fig. \ref{fig:epuck}) \cite{mondada2009epuck} in an indoor playground (Fig. \ref{fig:playground}). The task is called \emph{rendezvous}, where robots are randomly dispersed in the arena and must gather together.



We train these robots using the widely adopted Sim2Real paradigm \cite{hofer2021sim2real} in the RL community, in which agents first learn their policy in a simulated environment before being deployed in the real world with fixed parameters. To evaluate attack performance, we present results from both simulated and real-world scenarios in Fig. \ref{phys-result}. Each method was tested 10 times in the real world, leading to several key findings:


(1) AMI outperforms all baselines in both simulated and real-world environments. The improvement of AMI in the real world is statistically significant ($p<.05$) compared to all baselines under a paired samples t-test and, on average, 5.43 higher than the best-performing baseline in the real world.


(2) The superiority of AMI can be further demonstrated by agent behaviors. As shown in the final state photographed in Fig. \ref{phys-video}, under baseline attacks, victims gather together as usual without being influenced, with adversaries moving away (Fig. 5a-5d). However, with our AMI attack, one victim gets influenced by the adversary, and are fooled to gather with the adversary, instead of majority victims (Fig. \ref{subfig:phys-img-4}). Notably, our AMI is the only method to achieve this. \emph{See video demonstrations in \url{https://github.com/DIG-Beihang/AMI}.}

\subsection{AMI Attack in Simulation Environment}


Apart from real world results, we further evaluate the effectiveness of AMI in 12 tasks in simulated environments for completeness, including six discrete control tasks (SMAC) and six continuous control tasks (MAMujoco), demonstrating its superior performance. For simplicity, we assume the attacker controls the first agent in all environments. As shown in Fig. \ref{exp_all}, by strategically influencing victims toward a jointly worst target, our AMI outperforms the competing methods in 10 out of 12 environments across both continuous and discrete control, highlighting the effectiveness of our approach.

\subsection{Analyzation of AMI policies}

\begin{figure*}[t]
\centering
\includegraphics[scale=0.23]{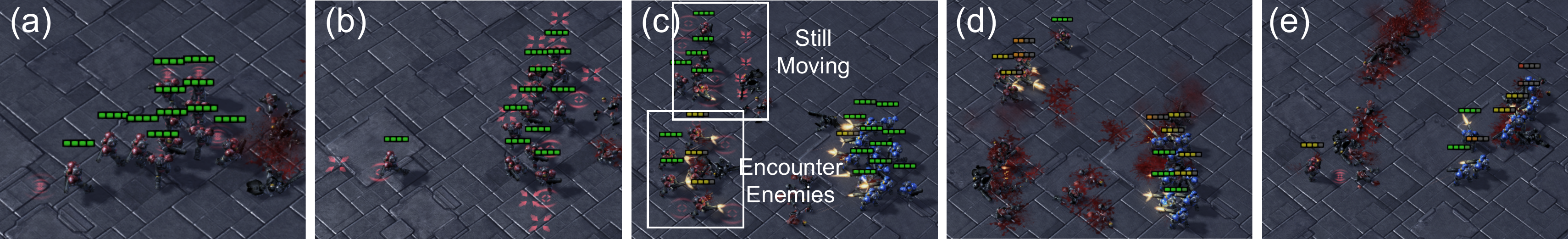}
\caption{Understanding the behavior of AMI. Attacker and victims in red, enemies in blue. (a) attackers entice victims to get back. (b) victims were influenced into a bad position. (c) some victims encounter enemies, while others are still moving. (d) first-arrived victims died, and enemies focused fire on the rest. (e) attacker moves near to get killed.}
\label{explaination-ami}
\vspace{-0.1in}
\end{figure*}

\begin{figure*}[htbp]
    \centering
    \captionsetup[subfloat]{font=tiny}
    \subfloat[TAO, Agent 3 (High AMI)]{
        \includegraphics[width=0.23\linewidth]{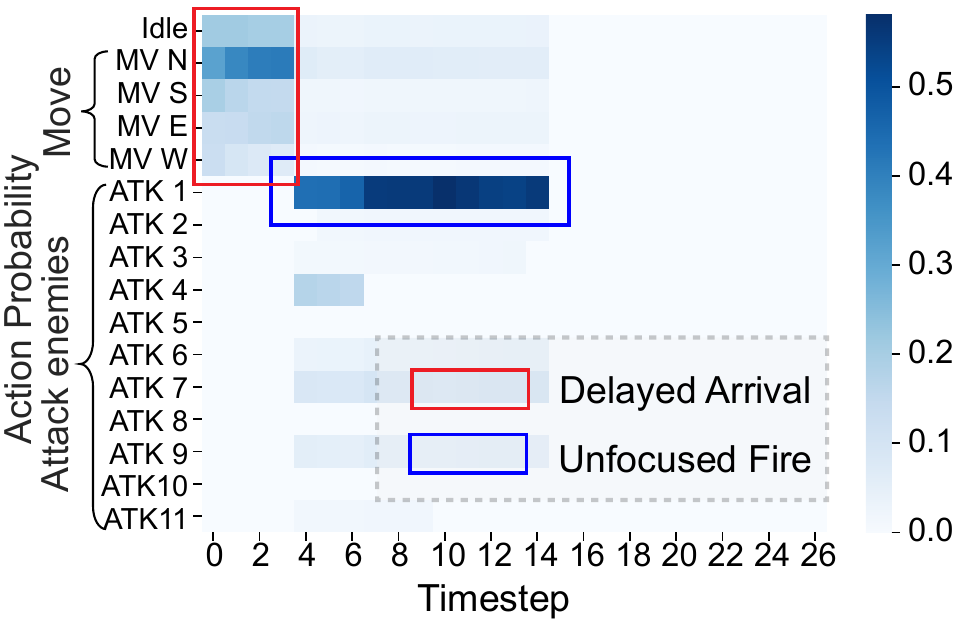}
        \label{subfig:exp-tao-1}
    }
    \subfloat[TAO, Agent 7 (Low AMI)]{
        \includegraphics[width=0.23\linewidth]{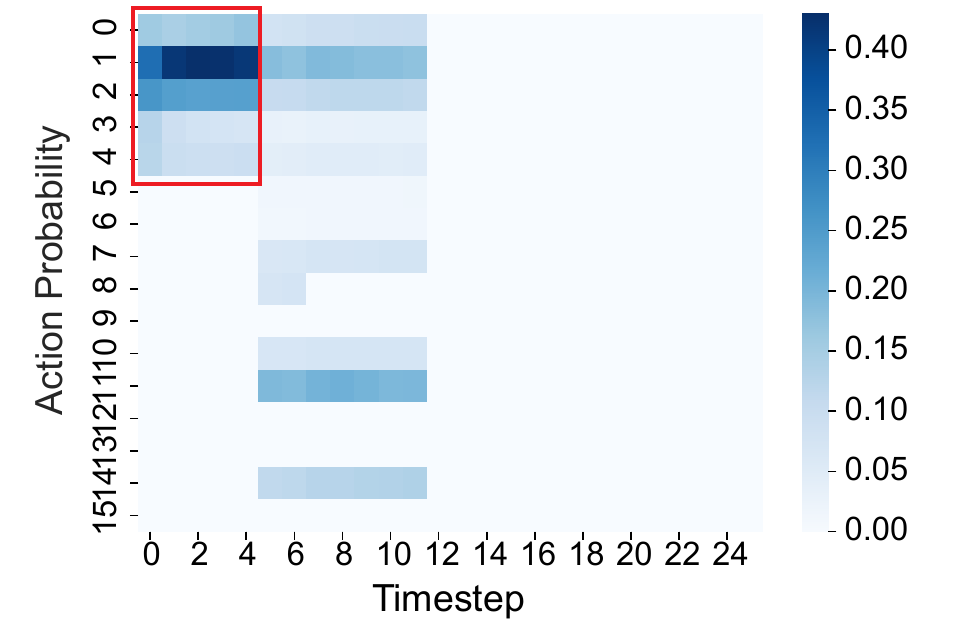}
        \label{subfig:exp-tao-2}
    }
    \subfloat[TAO, Agent 9 (High AMI)]{
        \includegraphics[width=0.23\linewidth]{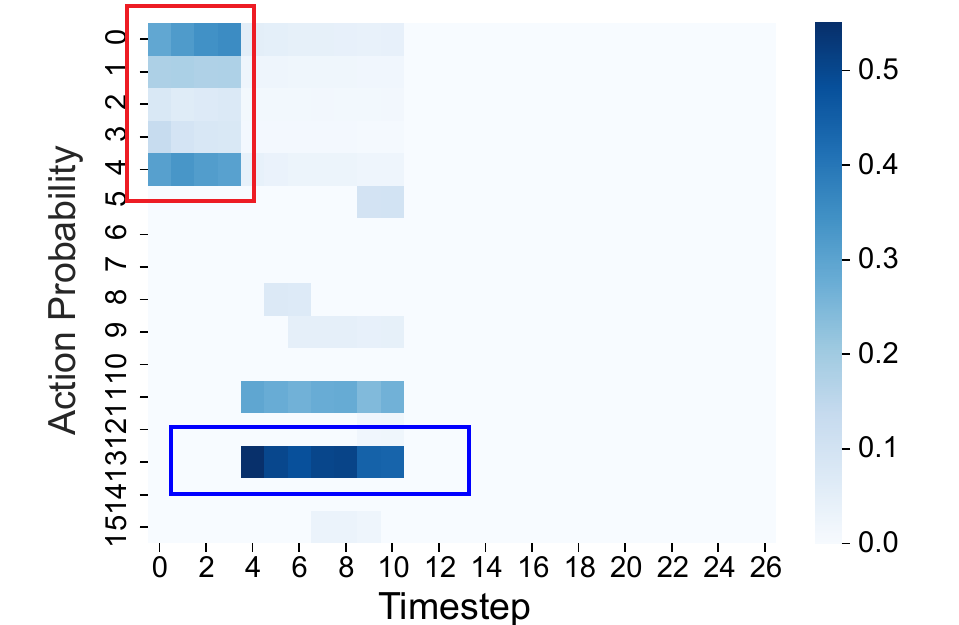}
        \label{subfig:exp-tao-3}
    }
    \subfloat[Joint Victim Actions]{
        \includegraphics[width=0.23\linewidth]{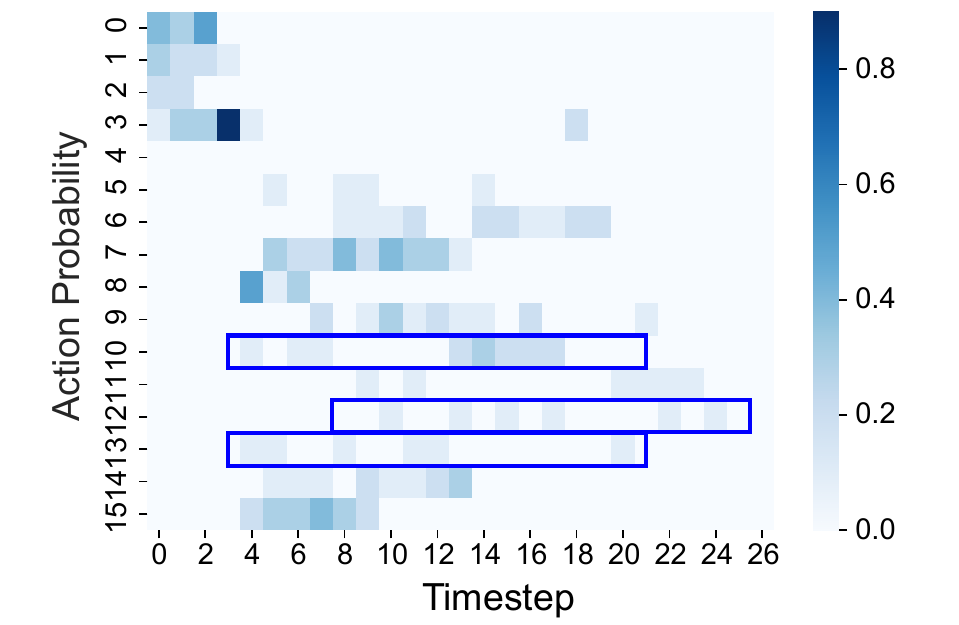}
        \label{subfig:exp-tao-4}
    }
    \caption{Actions suggested by TAO and taken by victims, evaluated in \emph{10m vs 11m}. Red and blue bracket indicates delayed arrival and unfocused fire behavior of AMI. MV N means move north and ATK X means attack enemy ID X.}
    \vspace{-0.2in}
    \label{explaination-tao}
\end{figure*}


To investigate the attack behavior of AMI, we visualize the behavior of victims subjected to AMI attacks in \emph{10m vs 11m} environment of SMAC, shown in Fig. \ref{explaination-ami} and \ref{explaination-tao}.


\textbf{Victim behavior under AMI.} Victims under AMI attack exhibit two critical behaviors that contributes to collective failure: (1) \emph{delayed arrival}. As depicted in Fig. \ref{explaination-ami}, victims are influenced by the attacker and divided into two groups. These groups encounter enemies at different timesteps: while some victims confront the full force of their adversaries, others are still approaching and do not have enemies within their firing range. (2) \emph{unfocused fire}. In SMAC, focused fire involves allies collaboratively attacking and defeating enemies one at a time. However, in the presence of the attacker, allies fire in an \emph{unfocused} manner. As illustrated in Fig. \ref{subfig:exp-tao-4}, victims' shots at enemies (Action ID 5-15) are randomly distributed, lacking a coordinated target. Consequently, victims fail to eliminate enemy units and face stronger enemy fire.


\textbf{Influence through the lens of TAO.} The behavior of victims under AMI can be explained by the target actions generated by TAO. As demonstrated in Fig. \ref{explaination-tao}, for understanding \emph{delayed arrival}, at the game's onset, victims 3 and 7 are encouraged to move north (Action ID 1), arriving later compared to agents moving directly east; for victim 9, which is moving east, it is encouraged to move west or stay idle: had it followed this target, it would have arrived slightly later than agents moving east directly (\ie, not being influenced), but still earlier than agents moving north, also resulting in delayed arrival. To comprehend \emph{unfocused fire}, agents 3, 7, and 9 are encouraged to attack \emph{different} enemies at the current timestep, limiting the number of victims attacking each enemy and suppressing focused fire.


\textbf{Adaptive target generation.} Furthermore, we discover that TAO can generate adaptive policies for victims with varying susceptibility, as illustrated in Fig. \ref{explaination-tao}. By calculating AMI, we find that agents 3 and 9 are more influenced, and their target policies learned by TAO are more deterministic. In this manner, TAO generates a \emph{targeted} goal for susceptible agents, as they have a higher probability of being influenced toward the collectively worst target policy. Conversely, since agent 7 is less influenced, the target policy learned by TAO is less deterministic. In this case, TAO generates an \emph{untargeted} goal for insusceptible agents: as their policies are difficult to influence, the attacker achieves the best results by preventing them from playing the optimal policy at the current timestep. In this way, TAO automatically learns different policies for susceptible and insusceptible victims under the current attacks.

\subsection{Ablations}

In this section, we verify the effectiveness of each component in our model. We conduct all experiments on \emph{3s5z vs 3s6z} for SMAC and \emph{HalfCheetah 6x1} for MAMujoco.

\subsubsection{Ablations on unilateral and targeted properties}

\begin{figure}[h]
    \centering
    \subfloat[SMAC environment]{
        \includegraphics[width=0.35\linewidth]{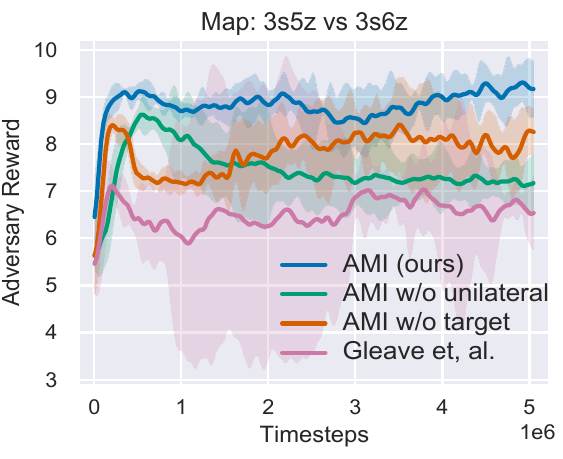}
        \label{fig:ablate-method-smac}
    }
    \subfloat[MAMujoco environment]{
        \includegraphics[width=0.35\linewidth]{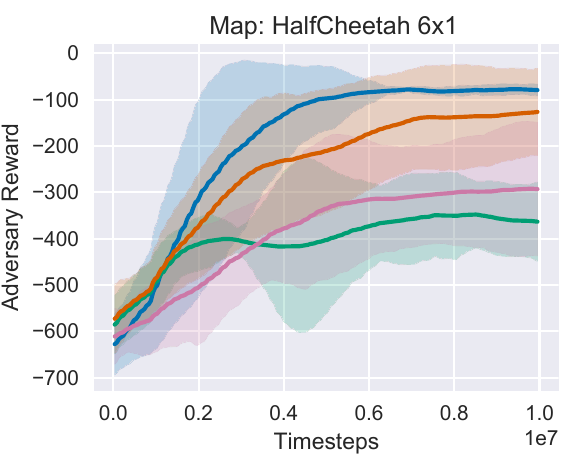}
        \label{fig:ablate-method-mujoco}
    }
 
    \caption{Ablation of unilateral and targeted properties of AMI. Both properties improve attack performance.}
    \vspace{-0.1in}
    \label{ablate-method}
\end{figure}


In this ablation study, we seek to assess the individual contributions of AMI's unilateral and targeted attack properties. For a version of AMI without unilateral attack capabilities, we integrate the majority influence term from Eqn. \ref{eqn:mutualinfo} into Eqn. \ref{eqn:advinf_final}, resulting in bilateral influence. In contrast, for AMI without a targeted attack, we employ Eqn. \ref{eqn:klrandom} as the influence metric. We also provide a comparison with the results of \emph{Gleave et al.} (\ie, absent of unilateral and targeted attacks). As illustrated in Fig. \ref{ablate-method}, the effectiveness of AMI relies on both enhancements. Notably, we observe that AMI without unilateral attack capabilities underperforms \emph{Gleave et al.} in the \emph{HalfCheetah 6x1} environment. This finding implies that majority influence can strongly diminish attack capability by encouraging the attacker to modify its actions to overfit to victim policies.

\subsubsection{Ablations on distance metric}

\begin{table}[h]
\centering
\small

\caption{In this section, we address the distance metrics employed for both discrete and continuous environments. AMI is determined using these metrics and subsequently maximized by the attacker under the optimal hyperparameter $\lambda$.}
\setlength\tabcolsep{10pt}
\label{distance_metric}
\begin{tabular}{ccc}
\hline
Name & Equation & Environment \\ \hline
  $\ell_1$      &  $-|| p(\hat{a}_{i}^\nu|s, \mathbf{a}) - \mathbf{1}_{\mathcal{A}}(a_{i}^\tau)||_{1}$ &  Discrete   \\
  $\ell_2$ &   $-|| p(\hat{a}_{i}^\nu|s, \mathbf{a}) - \mathbf{1}_{\mathcal{A}}(a_{i}^\tau)||_{2}$  &  Discrete     \\
 $\ell_\infty$   &  $-|| p(\hat{a}_{i}^\nu|s, \mathbf{a}) - \mathbf{1}_{\mathcal{A}}(a_{i}^\tau)||_{\infty}$ &  Discrete     \\
 CE &  $\log(p(a_{i}^\tau|s, \mathbf{a}))$  &   Discrete     \\
  Prob   &  $p(a_{i}^\tau|s, \mathbf{a})$   &   Discrete      \\ \hline
  $\ell_1$ &  $-|| \hat{\mu}_{i}^\nu(s, \mathbf{a}) - a_{i}^\tau||_{1}$   &    Continuous  \\
  CE   & $\log(p(a_{i}^\tau|s, \mathbf{a}))$  &   Continuous    \\
  Prob   & $p(a_{i}^\tau|s, \mathbf{a})$  &  Continuous \\ \hline
\end{tabular}
\vspace{-0.1in}
\end{table}

\begin{figure}[h]
    \centering
    \subfloat[Distance - Discrete]{
        \includegraphics[width=0.35\linewidth]{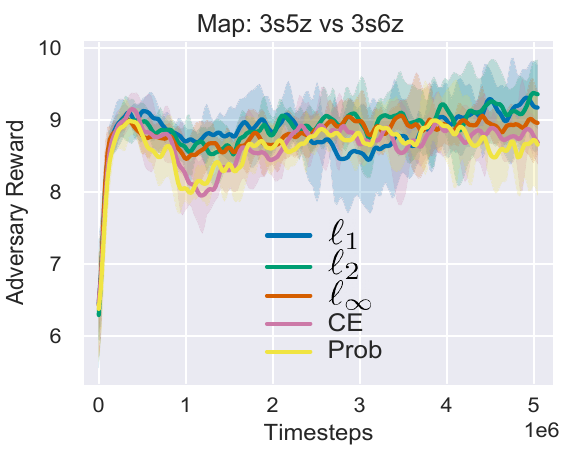}
        \label{fig:distance-discrete}
    }
    \subfloat[Distance - Continuous]{
        \includegraphics[width=0.35\linewidth]{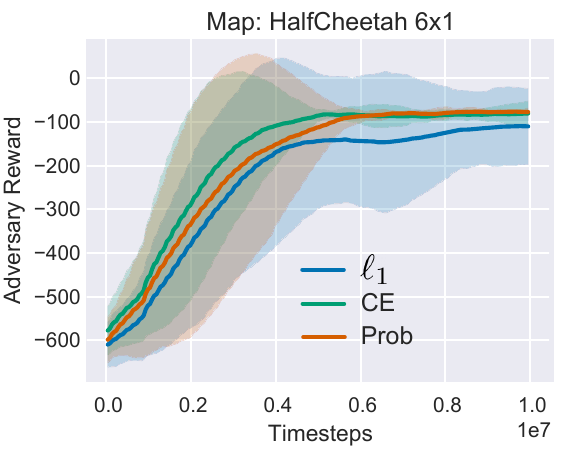}
        \label{fig:distance-continuous}
    }
    \caption{Ablation on distance metric. The performance of AMI is stable under different distance metrics.}
    \vspace{-0.2in}
    \label{ablate-distance}
\end{figure}


Next, we investigate the performance of AMI when utilizing various distance metrics. We present alternative distance metrics for both discrete and continuous environments in Table \ref{distance_metric}, where $\hat{\mu}_{i}^\nu(s, \mathbf{a})$ represents the mean predicted by the opponent modeling for continuous control, under the assumption that actions adhere to a Gaussian distribution. The outcomes are illustrated in Fig. \ref{ablate-distance}, which reveals that the attack capability of AMI remains largely unaffected by the choice of distance metrics for both continuous and discrete control scenarios. Notably, the metrics employed in our AMI ($\ell_1$ for discrete control and Prob for continuous control) yield the most favorable results.

\subsubsection{Comparing with mutual information}
\begin{figure}[h]
    \centering
    \vspace{-0.1in}
    \subfloat[SMAC environment]{
        \includegraphics[width=0.35\linewidth]{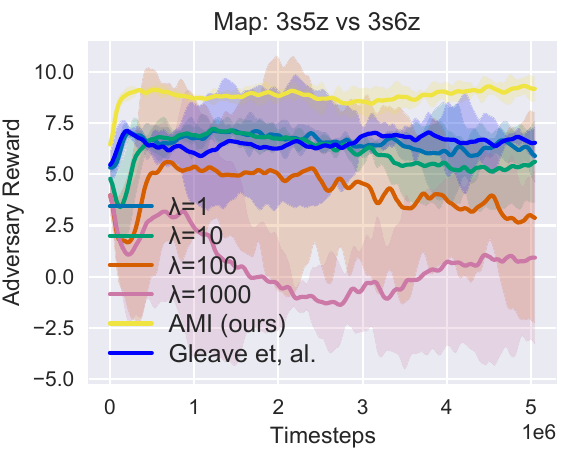}
        \label{fig:mi-method-smac}
    }
    \subfloat[MAMujoco environment]{
        \includegraphics[width=0.35\linewidth]{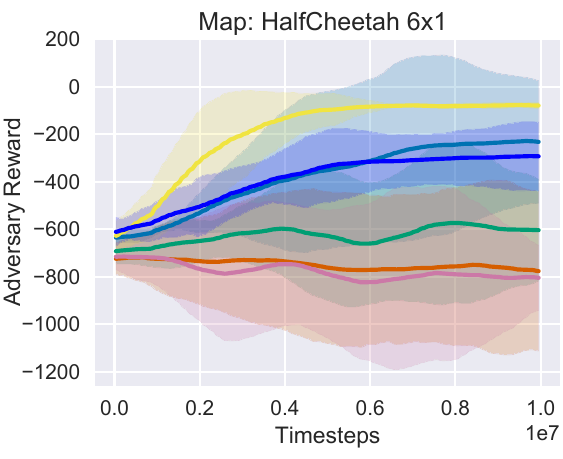}
        \label{fig:mi-method-mujoco}
    }
    \caption{Using mutual information proposed in \cite{jaques2019icmlsocialinfluence} as influence metric. Since mutual information is bidirectional, such metric was ineffective for attack.}
    \vspace{-0.1in}
    \label{discussion-mi}
\end{figure}

Since unilateral influence is derived from mutual information, we also take mutual information \cite{jaques2019icmlsocialinfluence} that depicts the bilateral influence between victims and adversary as a baseline. Specifically, we evaluate the result for four different $\lambda$ values that modulate the magnitude of mutual information. As illustrated in Fig. \ref{discussion-mi}, utilizing mutual information often results in inferior performance compared to \emph{Gleave et al.} (\ie, not employing mutual information). Moreover, the outcome worsens as $\lambda$ increases. These observations serve as motivation for the development of AMI, a unilateral and targeted influence designed for c-MARL attacks, which yields superior performance.

\subsubsection{Ablations on hyperparameter}

\begin{figure}[h]
    \centering
    \vspace{-0.1in}
    \subfloat[SMAC environment]{
        \includegraphics[width=0.35\linewidth]{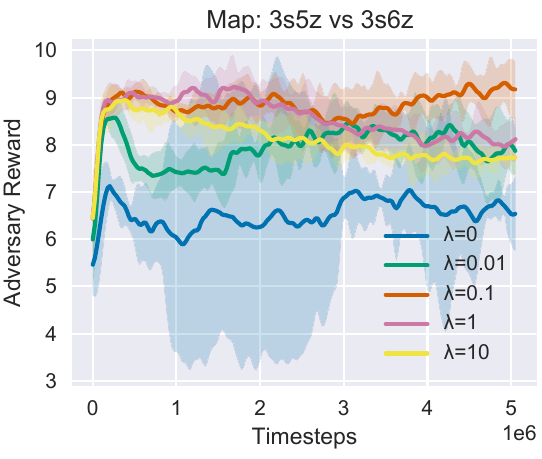}
        \label{fig:ablate-hyperparam-smac}
    }
    \subfloat[MAMujoco environment]{
        \includegraphics[width=0.35\linewidth]{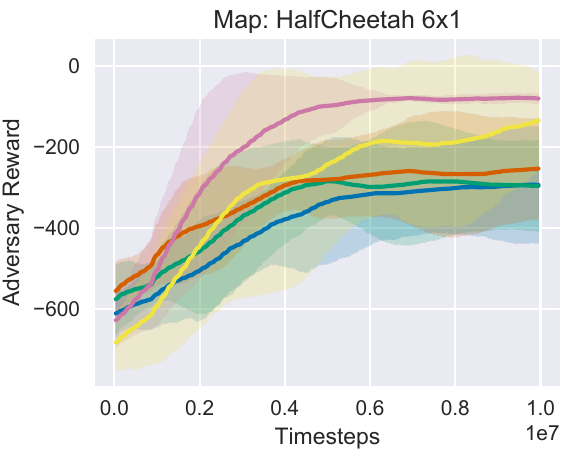}
        \label{fig:ablate-hyperparam-mujoco}
    }

    \caption{Ablation on hyperparameter $\lambda$.}
    \vspace{-0.1in}
    \label{ablate-hyperparam}
\end{figure}


Lastly, we evaluate AMI under varying hyperparameters $\lambda$, which modulate the contribution of the AMI reward $I_t^\alpha$ to the total reward. Specifically, we assess our AMI attack for $\lambda = \{0, 0.01, 0.1, 1, 10\}$, where $\lambda = 0$ reduces to the method of \emph{Gleave et al}. As depicted in Fig. \ref{ablate-hyperparam}, incorporating AMI enhances the overall attack performance compared to \emph{Gleave et al}, thereby demonstrating the effectiveness of our approach. Moreover, selecting an appropriate hyperparameter leads to optimal AMI performance ($\lambda=0.1$ in \emph{3s5z vs 3s6z} and $\lambda=1$ in \emph{HalfCheetah 6x1}). We also observe that performance does not improve monotonically with increasing $\lambda$ values. This can be attributed to the following factors: (1) a large $\lambda$ generates high rewards, which may introduce instability when training the critic network, and (2) a substantial $\lambda$ causes errors in the opponent modeling module $p_\phi$ to accumulate, resulting in excessive emphasis on the attack capability of the subsequent timestep.

\section{Conclusion}


In this paper, we present AMI as a strong and practical black-box attack to assess the robustness of c-MARL, in which the attacker unilaterally influences victims to establish a worst-case collaboration. Firstly, we adapt agent-wise bilateral mutual information to a unilateral adversary-victim influence for policy-based attacks by decomposing mutual information and filtering out the influence from victims to the adversary. Secondly, we employ a reinforcement learning agent to generate the jointly worst-case target for the attacker to influence in order to maximize team reward. Through AMI, we pioneers the successful execution of adversarial attacks on real-world robotic swarms, and demonstrate its effectiveness in compelling agents towards collectively unfavorable outcomes in simulated environments. Our findings not only offer valuable insights into system vulnerabilities but also open opportunities to strengthen the resilience of cooperative multi-agent systems.



\

\textbf{Acknowledgements}

This work was supported by the National Key Research and Development Plan of China (2022ZD0116405), the National Natural Science Foundation of China (62022009, 62206009, 62306025), and the State Key Laboratory of Software Development Environment.

\appendix

\section{Detailed Derivation of Eqn. 5}
\label{appendix_proofa}
Here we present the detailed derivation of Eqn. 5 in our main paper.

\begin{equation}
\begin{split}
\label{eqn:klrandom}
&H(\hat{a}_{t+1, i}^\nu|s_t, \mathbf{a}_t^\nu) \\
& = - \sum_{\hat{a}_{t+1, i}^\nu \in \mathcal{A}_i} p_\phi(\hat{a}_{t+1, i}^\nu|s_t, \mathbf{a}_t^\nu) \log\left(p_\phi(\hat{a}_{t+1, i}^\nu|s_t, \mathbf{a}_t^\nu)\right) \\
& = - \sum_{\hat{a}_{t+1, i} \in \mathcal{A}_i} p_\phi(\hat{a}_{t+1, i}^\nu|s_t, \mathbf{a}_t^\nu) \log\left(p_\phi(\hat{a}_{t+1, i}^\nu|s_t, \mathbf{a}_t^\nu\right) - p_\phi(\hat{a}_{t+1, i}^\nu|s_t, \mathbf{a}_t^\nu) \log(\mathcal{U}) + c,  \\
& = - D_{KL}\left(p_\phi(\hat{a}_{t+1, i}^\nu|s_t, \mathbf{a}_t^\nu)|| \ \mathcal{U})\right) + c \\
& = - D_{KL}(\left[\sum_{\tilde{a}_t^\alpha} p_\phi(\hat{a}_{t+1, i}^\nu|s_t, \tilde{a}_t^\alpha, \mathbf{a}_t^\nu)\cdot p(\tilde{a}_t^\alpha|s_t, \mathbf{a}_t^\nu)\right]|| \ \mathcal{U})) + c, \\
& = - D_{KL}(\left[\sum_{\tilde{a}_t^\alpha} p_\phi(\hat{a}_{t+1, i}^\nu|s_t, \tilde{a}_t^\alpha, \mathbf{a}_t^\nu)\cdot \pi^\alpha(\tilde{a}_t^\alpha|s_t)\right]|| \ \mathcal{U})) + c, \\
&= - D_{KL}\left(\mathbb{E}_{\tilde{a}_t^\alpha \sim \pi^\alpha}\left[p_\phi(\hat{a}_{t+1, i}^\nu|s_t, \tilde{a}_t^\alpha, \mathbf{a}_t^\nu)\right]|| \ \mathcal{U})\right) + c,
\end{split}
\end{equation}

\section{Convergence of Value Function $Q^{\tau}(s, a^{\tau}, a^\alpha)$}
\label{appendix_proofb}
Here we present the full proof of convergence of $Q^{\tau}(s, a^{\tau}, a^\alpha)$. We first show that updating $Q^{\tau}$ by Bellman operator $\mathcal{B}^{\tau}$ is a contraction on Banach space, with $\mathcal{B}^{\tau}$ defined as:
\begin{equation}
\label{eqn:tao_bellman_operator}
\begin{split}
\left(\mathcal{B}^{\tau} Q^{\tau}\right)(s, a^{\tau}, a^\alpha) =  & r_t^\alpha + \gamma \sum_{s'} \mathcal{T}(s'|s, a^\alpha, \mathbf{a}^\nu) \sum_{a'^\tau \in \mathcal A} \\
& \pi(a'^\tau|s') \hspace{-18pt}\sum_{\hspace{18pt}(\mathbf{a}'^\nu, a'^\alpha) \in \mathcal A} \hspace{-18pt} \pi^\alpha(a'^\alpha|h_i') \pi^\nu(\mathbf{a}'^\alpha|\mathbf{h}') Q^\tau(s', a'^{\tau}, a'^\alpha).
\end{split}
\end{equation}

Define two Q functions $Q^{\tau}_1(s, a^{\tau}, a^\alpha)$ and $Q^{\tau}_2(s, a^{\tau}, a^\alpha)$, we need to show the Bellman operator $\mathcal{B}^{\tau}$ is a contraction in sup-norm:

\begin{equation}
\label{eqn:tao_bellman_operator}
\begin{split}
& \quad || \mathcal{B}^{\tau} Q^{\tau}_1 - \mathcal{B}^{\tau} Q^{\tau}_2 ||_\infty \\
&=  \max_{s, a^\tau, a^\alpha}| (\mathcal{B}^{\tau} Q^{\tau}_1)(s, a^{\tau}, a^\alpha) - (\mathcal{B}^{\tau} Q^{\tau}_2)(s, a^{\tau}, a^\alpha) | \\
&=  \max_{s, a^\tau, a^\alpha} \Bigg| r_t^\alpha + \gamma \mathcal{T}(s'|s, a^\alpha, \mathbf{a}^\nu) \sum_{a'^\tau \in \mathcal A} \pi(a'^\tau|s') \hspace{-18pt}\sum_{\hspace{18pt}(\mathbf{a}'^\nu, a'^\alpha) \in \mathcal A} \hspace{-18pt} \\ 
& \quad \quad \pi^\alpha(a'^\alpha|h'_i) \pi^\nu(\mathbf{a}'^\alpha|\mathbf{h}') Q^\tau_1(s', a'^{\tau}, a'^\alpha) - r_t^\alpha - \gamma \mathcal{T}(s'|s, a^\alpha, \mathbf{a}^\nu) \\
& \quad \quad \sum_{a'^\tau \in \mathcal A} \pi(a'^\tau|s') \hspace{-18pt}\sum_{\hspace{18pt}(\mathbf{a}'^\nu, a'^\alpha) \in \mathcal A} \hspace{-18pt} \pi^\alpha(a'^\alpha|h_i') \pi^\nu(\mathbf{a}'^\alpha|\mathbf{h}') Q^\tau_2(s', a'^{\tau}, a'^\alpha) \Bigg| \\
&=  \max_{s, a^\tau, a^\alpha} \gamma \Bigg| \mathcal{T}(s'|s, a^\alpha, \mathbf{a}^\nu) \sum_{a'^\tau \in \mathcal A} \pi(a'^\tau|s') \hspace{-18pt}\sum_{\hspace{18pt}(\mathbf{a}'^\nu, a'^\alpha) \in \mathcal A} \hspace{-18pt}\pi^\alpha(a'^\alpha|h_i') \\
& \quad \quad \pi^\nu(\mathbf{a}'^\alpha|\mathbf{h}') (Q^\tau_1(s', a'^{\tau}, a'^\alpha) - Q^\tau_2(s', a'^{\tau}, a'^\alpha)) \Bigg| \\
&\leq  \max_{s, a^\tau, a^\alpha} \gamma \mathcal{T}(s'|s, a^\alpha, \mathbf{a}^\nu) \sum_{a'^\tau \in \mathcal A} \pi(a'^\tau|s') \hspace{-18pt}\sum_{\hspace{18pt}(\mathbf{a}'^\nu, a'^\alpha) \in \mathcal A} \hspace{-18pt}\pi^\alpha(a'^\alpha|h_i') \\
& \quad \quad \pi^\nu(\mathbf{a}'^\alpha|\mathbf{h}') |Q^\tau_1(s', a'^{\tau}, a'^\alpha) - Q^\tau_2(s', a'^{\tau}, a'^\alpha)| \\
&\leq  \gamma ||Q^\tau_1(s', a'^{\tau}, a'^\alpha) - Q^\tau_2(s', a'^{\tau}, a'^\alpha)||_\infty
\end{split}
\end{equation}

Thus, $\mathcal{B}^{\tau}$ is a contraction operator. Finally, by Banach's fixed point theorem, with finite joint action space $\mathcal A$, state space $\mathcal S$, and assume each state-action pair is visited infinitesimally often, updating $Q^{\tau}(s, a^{\tau}, a^\alpha)$ by Bellman operator $\mathcal{B}^{\tau}$ will converge to the optimal value function $Q^{\tau, *}(s, a^{\tau}, a^\alpha)$. Note that the guaranteed convergence happens in tabular case. This motivates us to use PPO algorithm as a practical solver of this problem.

\section{Convergence of Value Function $Q^{\alpha}(s, a^{\tau}, a^\alpha)$}
\label{appendix_proofc}
The convergence of $Q^{\alpha}(s, a^{\alpha}, a^\alpha)$ follows the same technique with the convergence proof of $Q^{\tau}(s, a^{\tau}, a^\alpha)$. We state the proof here again for completeness.

Again, we first show that updating $Q^{\alpha}$ by Bellman operator $\mathcal{B}^{\alpha}$ is a contraction on Banach space, with $\mathcal{B}^{\alpha}$ defined as:
\begin{equation}
\label{eqn:adv_bellman_operator}
\begin{split}
\left(\mathcal{B}^\alpha Q^{\alpha}\right)(s, a^{\tau}, a^\alpha) = & r^\alpha + \lambda \cdot I^\alpha+ \gamma \sum_{s'} \mathcal{T}(s'|s, a^\alpha, \mathbf{a}^\nu) \\
& \sum_{a'^\tau \in \mathcal A} \pi(a'^\tau|s') \hspace{-18pt}\sum_{\hspace{18pt}(\mathbf{a}'^\nu, a'^\alpha) \in \mathcal A} \hspace{-18pt} \pi^\alpha(a'^\alpha|h_i') \pi^\nu(\mathbf{a}'^\alpha|\mathbf{h}') Q^\alpha(s', a'^{\tau}, a'^\alpha).
\end{split}
\end{equation}

Define two Q functions $Q^{\alpha}_1(s, a^{\tau}, a^\alpha)$ and $Q^{\alpha}_2(s, a^{\tau}, a^\alpha)$, we need to show the Bellman operator $\mathcal{B}^{\alpha}$ is a contraction in sup-norm:

\begin{equation}
\label{eqn:tao_bellman_operator}
\begin{split}
& \quad || \mathcal{B}^{\alpha} Q^{\alpha}_1 - \mathcal{B}^{\alpha} Q^{\alpha}_2 ||_\infty \\
& =  \max_{s, a^\tau, a^\alpha}| (\mathcal{B}^{\alpha} Q^{\alpha}_1)(s, a^{\tau}, a^\alpha) - (\mathcal{B}^{\alpha} Q^{\alpha}_2)(s, a^{\tau}, a^\alpha) | \\
&=  \max_{s, a^\tau, a^\alpha} \Bigg| r_t^\alpha + \lambda \cdot I^\alpha + \gamma \mathcal{T}(s'|s, a^\alpha, \mathbf{a}^\nu) \sum_{a'^\tau \in \mathcal A} \pi(a'^\tau|s')  \\ 
& \quad \quad \hspace{-18pt}\sum_{\hspace{18pt}(\mathbf{a}'^\nu, a'^\alpha) \in \mathcal A} \hspace{-18pt} \pi^\alpha(a'^\alpha|h'_i) \pi^\nu(\mathbf{a}'^\alpha|\mathbf{h}') Q^\alpha_1(s', a'^{\tau}, a'^\alpha) - r_t^\alpha \\
& \quad \quad - \lambda \cdot I^\alpha - \gamma \mathcal{T}(s'|s, a^\alpha, \mathbf{a}^\nu) \sum_{a'^\tau \in \mathcal A} \pi(a'^\tau|s') \\
& \quad \quad \hspace{-18pt}\sum_{\hspace{18pt}(\mathbf{a}'^\nu, a'^\alpha) \in \mathcal A} \hspace{-18pt} \pi^\alpha(a'^\alpha|h_i') \pi^\nu(\mathbf{a}'^\alpha|\mathbf{h}') Q^\alpha_2(s', a'^{\tau}, a'^\alpha) \Bigg| \\
&=  \max_{s, a^\tau, a^\alpha} \gamma \Bigg| \mathcal{T}(s'|s, a^\alpha, \mathbf{a}^\nu) \sum_{a'^\tau \in \mathcal A} \pi(a'^\tau|s') \hspace{-18pt}\sum_{\hspace{18pt}(\mathbf{a}'^\nu, a'^\alpha) \in \mathcal A} \hspace{-18pt}\pi^\alpha(a'^\alpha|h_i') \\
& \quad \quad \pi^\nu(\mathbf{a}'^\alpha|\mathbf{h}') (Q^\alpha_1(s', a'^{\tau}, a'^\alpha) - Q^\alpha_2(s', a'^{\tau}, a'^\alpha)) \Bigg| \\
&\leq  \max_{s, a^\tau, a^\alpha} \gamma \mathcal{T}(s'|s, a^\alpha, \mathbf{a}^\nu) \sum_{a'^\tau \in \mathcal A} \pi(a'^\tau|s') \hspace{-18pt}\sum_{\hspace{18pt}(\mathbf{a}'^\nu, a'^\alpha) \in \mathcal A} \hspace{-18pt}\pi^\alpha(a'^\alpha|h_i') \\
& \quad \quad \pi^\nu(\mathbf{a}'^\alpha|\mathbf{h}') |Q^\alpha_1(s', a'^{\tau}, a'^\alpha) - Q^\alpha_2(s', a'^{\tau}, a'^\alpha)| \\
 & \leq \gamma ||Q^\alpha_1(s', a'^{\tau}, a'^\alpha) - Q^\alpha_2(s', a'^{\tau}, a'^\alpha)||_\infty
\end{split}
\end{equation}

Thus, $\mathcal{B}^{\alpha}$ is a contraction operator. Finally, by Banach's fixed point theorem, with finite joint action space $\mathcal A$, state space $\mathcal S$, and assume each state-action pair is visited infinitesimally often, updating $Q^{\alpha}(s, a^{\tau}, a^\alpha)$ by Bellman operator $\mathcal{B}^{\alpha}$ will converge to the optimal value function $Q^{\alpha, *}(s, a^{\tau}, a^\alpha)$. Again, the guaranteed convergence happens in tabular case. We thus use PPO algorithm as a practical solver of this problem.

\section{Experiment Hyperparameters}
\label{appendix_hyperparam}
In this section, we describe the hyperparameters of AMI and baselines. For \textbf{SMAC environment}, Table. \ref{params-smac-share} describes the shared parameters used by all methods in SMAC experiments. Table. \ref{params-smac-method} denotes the hyperparameters used for each individual methods.

\begin{table}[htbp]
\centering
\small
\caption{Shared hyperparameters for SMAC, used in AMI and all baselines.}
\vspace{+0.1in}
\label{params-smac-share}
\begin{tabular}{cc|cc}
\hline
Hyperparameter & Value & Hyperparameter & Value \\ \hline
       lr      &  1e-4 &  mini-batch num &   1  \\
 parallel envs &   32  &  max grad norm  &   10    \\
     gamma     &  0.99 &  max episode len &  150     \\
 actor network &  mlp  &   actor lr       &  =lr    \\
  hidden dim   &  64   &   critic lr      &  =lr     \\
  hidden layer &  1   &    PPO epoch      &    4   \\
  activation   & ReLU  &   PPO clip    &   0.2    \\
   optimizer   & Adam  &  entropy coef   &  0.01   \\
  GAE lambda   & 0.95  &  eval episode   &  20   \\ \hline
\end{tabular}
\end{table}

\begin{table}[htbp]
\centering
\small
\caption{Method-specific parameters for \emph{Wu et al.}, \emph{Guo et al.} and our AMI in SMAC environment.}
\vspace{+0.1in}
\label{params-smac-method}
\begin{tabular}{cccc}
\hline
\multicolumn{4}{c}{Hyperparameters for Wu et al.}                                             \\ \hline
Hyperparameter            & \multicolumn{1}{c|}{Value} & Hyperparameter  & Value              \\ \hline
epsilon\_state            & \multicolumn{1}{c|}{0.1}   & epsilon\_action & 0.1                \\ 
lr\_state            & \multicolumn{1}{c|}{=actor lr}   & lr\_action & =actor lr                \\ \hline
\multicolumn{4}{c}{Hyperparameters for \emph{Guo et al.}}                                            \\ \hline
Hyperparameter            & \multicolumn{1}{c|}{Value} & Hyperparameter  & Value              \\ \hline
victim critic lr          & \multicolumn{1}{c|}{=lr}   &                 &                    \\ \hline
\multicolumn{4}{c}{Hyperparameters for AMI}                                                   \\ \hline
Hyperparameter            & \multicolumn{1}{c|}{Value} & Hyperparameter  & Value              \\ \hline
$p_\phi$ lr & \multicolumn{1}{c|}{=lr}   & TAO lr          & =lr                \\
TAO critic lr             & \multicolumn{1}{c|}{=lr}   & AMI lambda      & [.03, .05, .1] \\ \hline
\end{tabular}
\end{table}

For \textbf{MAMujoco environment}, Table. \ref{params-mujoco-all} describes the parameters used for all experiments. Table. \ref{params-mujoco-method} denotes the hyperparameters used for each individual methods. 

\begin{table}[htbp]
\centering
\caption{Shared hyperparameters for MAMujoco, used in AMI and all baselines in MAMujoco environment.}
\vspace{+0.1in}
\label{params-mujoco-all}
\begin{tabular}{cc|cc}
\hline
Hyperparameter & Value & Hyperparameter & Value \\ \hline
parallel envs  &  32 &  mini-batch num &   40  \\
 gamma &  0.99   &  max grad norm  &   10    \\
 gain  &  0.01 &  max episode len &  1000     \\
 actor network &  mlp  &   actor lr  &  [5e-6, 5e-4]    \\
 std y coef &  0.5  &   critic lr    &  5e-3    \\
 std x coef &  1  &   Huber loss   &  True    \\
  hidden dim   &  64   &   Huber delta    &   10    \\
  hidden layer &  1   &    PPO epoch      &    5   \\
  activation   & ReLU  &   PPO clip    &   0.2    \\
   optimizer   & Adam  &  entropy coef   &  0.01   \\ 
  GAE lambda   & 0.95  &  eval episode   &  32 \\ \hline
\end{tabular}
\end{table}

\begin{table}[htbp]
\centering
\caption{Method-specific parameters for \emph{Wu et al.}, \emph{Guo et al.} and our AMI in MAMujoco environment.}
\vspace{+0.1in}
\small
\label{params-mujoco-method}
\begin{tabular}{cccc}
\hline
\multicolumn{4}{c}{Hyperparameters for Wu et al.}                                             \\ \hline
Hyperparameter            & \multicolumn{1}{c|}{Value} & Hyperparameter  & Value              \\ \hline
epsilon\_state            & \multicolumn{1}{c|}{0.1}   & epsilon\_action & 0.1                \\
lr\_state            & \multicolumn{1}{c|}{=actor lr}   & lr\_action & =actor lr                \\ \hline
\multicolumn{4}{c}{Hyperparameters for \emph{Guo et al.}}                                            \\ \hline
Hyperparameter            & \multicolumn{1}{c|}{Value} & Hyperparameter  & Value              \\ \hline
victim critic lr          & \multicolumn{1}{c|}{=lr}   &                 &                    \\ \hline
iteration             & \multicolumn{1}{c|}{30}   &  &   
\\ \hline
\multicolumn{4}{c}{Hyperparameters for AMI}                                                   \\ \hline
Hyperparameter            & \multicolumn{1}{c|}{Value} & Hyperparameter  & Value              \\ \hline
$p_\phi$ lr & \multicolumn{1}{c|}{=lr}   & TAO lr          & =lr                \\
TAO critic lr             & \multicolumn{1}{c|}{=lr}   & AMI lambda      & [.01, .1, .3, 1] \\ \hline
\end{tabular}
\end{table}

For \textbf{rendezvous environment}, the hyperparameters and implementations follows MAMujoco environment, with small variations to achieve best performance in this scenario. Table. \ref{params-juji-all} describes the parameters used for the rendezvous environment. Table. \ref{juji-method} denotes the hyperparameters used for each individual methods.

\begin{table}[htbp]
\centering
\small
\caption{Shared hyperparameters for rendezvous, used in AMI and all baselines.}
\vspace{+0.1in}
\label{params-juji-all}
\begin{tabular}{cc|cc}
\hline
Hyperparameter & Value & Hyperparameter & Value \\ \hline
parallel envs  &  32 &  mini-batch num &   40  \\
 gamma &  0.99   &  max grad norm  &   10    \\
 gain  &  0.01 &  max episode len &  1000     \\
 actor network &  mlp  &   actor lr  &  5e-5    \\
 std y coef &  0.5  &   critic lr    &  5e-3    \\
 std x coef &  1  &   Huber loss   &  True    \\
  hidden dim   &  64   &   Huber delta    &   10    \\
  hidden layer &  1   &    PPO epoch      &    5   \\
  activation   & ReLU  &   PPO clip    &   0.2    \\
   optimizer   & Adam  &  entropy coef   &  0.01   \\ 
  GAE lambda   & 0.95  &  eval episode   &  32 \\ \hline
\end{tabular}
\end{table}

\begin{table}[htbp]
\centering
\small
\caption{Method-specific parameters for \emph{Wu et al.}, \emph{Guo et al.} and our AMI in rendezvous environment.}
\vspace{+0.1in}
\label{juji-method}
\begin{tabular}{cccc}
\hline
\multicolumn{4}{c}{Hyperparameters for Wu et al.}                                             \\ \hline
Hyperparameter            & \multicolumn{1}{c|}{Value} & Hyperparameter  & Value              \\ \hline
epsilon\_state            & \multicolumn{1}{c|}{0.1}   & epsilon\_action & 0.1                \\
lr\_state            & \multicolumn{1}{c|}{=actor lr}   & lr\_action & =actor lr                \\ \hline
\multicolumn{4}{c}{Hyperparameters for \emph{Guo et al.}}                                            \\ \hline
Hyperparameter            & \multicolumn{1}{c|}{Value} & Hyperparameter  & Value              \\ \hline
victim critic lr          & \multicolumn{1}{c|}{=lr}   &                 &                    \\ \hline
\multicolumn{4}{c}{Hyperparameters for AMI}                                                   \\ \hline
Hyperparameter            & \multicolumn{1}{c|}{Value} & Hyperparameter  & Value              \\ \hline
$p_\phi$ lr & \multicolumn{1}{c|}{=lr}   & TAO lr          & =lr                \\
TAO critic lr             & \multicolumn{1}{c|}{=lr}   & AMI lambda      & .003 \\ \hline
\end{tabular}
\end{table}

\bibliographystyle{elsarticle-num}
\bibliography{ref.bib}

\end{document}